%% file: main_long.tex
\pdfoutput=1

\documentclass[11pt,usenames,dvipsnames]{article}

\usepackage{acl}

\usepackage{xcolor}
\usepackage{times}
\usepackage{tcolorbox}
\usepackage{latexsym}
\usepackage{graphicx}
\usepackage[T1]{fontenc}
\usepackage[utf8]{inputenc}
\usepackage{subfigure}
\usepackage{subcaption}
\usepackage{url}
\usepackage{microtype}
\usepackage{enumitem}
\usepackage{inconsolata}
\usepackage{amsfonts}
\usepackage{multirow}
\usepackage{float}
\usepackage{CJKutf8}
\usepackage{amsmath}
\usepackage{stmaryrd}
\usepackage{diagbox}
\usepackage{booktabs}
\usepackage{makecell}
\usepackage{svg}
\usepackage{ulem}
\usepackage{tabularx}
\usepackage{array}
\usepackage{wrapfig}

\usepackage[T1]{fontenc}

\usepackage[utf8]{inputenc}

\usepackage{microtype}

\usepackage{inconsolata}
\DeclareUnicodeCharacter{2212}{-}

\newcommand{\STAB}[1]{\begin{tabular}{@{}c@{}}#1\end{tabular}}
\newcommand{\gptmodel}[0]{\text{GPT-3.5}}
\newcommand{\dataset}[0]{\text{S-MedQA}}

\definecolor{niceblue}{HTML}{072AC8}
\definecolor{niceorange}{HTML}{F86624}
\newcommand\green[1]{\textcolor{niceblue}{#1}}
\newcommand\red[1]{\textcolor{niceorange}{#1}}

\title{What Does \textit{Neuro} Mean to \textit{Cardio}? \\ Investigating the Role of Clinical Specialty Data in Medical LLMs}

\author{
    Xinlan Yan\textsuperscript{1,2} \quad
    Di Wu\textsuperscript{2}\thanks{\; Corresponding author.} \quad Yibin Lei\textsuperscript{2} \quad Christof Monz\textsuperscript{2} \quad
    Iacer Calixto\textsuperscript{1,2}\\
    \textsuperscript{1}Department of Medical Informatics, Amsterdam UMC\\
    \textsuperscript{2}University of Amsterdam\\
    \texttt{x.yan@amsterdamumc.nl, d.wu@uva.nl, y.lei@uva.nl,}\\
    \texttt{c.monz@uva.nl, 
    i.coimbra@amsterdamumc.nl}
}

\begin{document}
\maketitle

\begin{abstract}
In this paper, we introduce \dataset{}, an English medical question-answering (QA) dataset for benchmarking large language models (LLMs) in fine-grained clinical specialties.
\dataset{} has over $24$k examples, covers $15$ medical specialties, and QA pairs can have multiple specialty annotations, e.g., when a question is cross-disciplinary, constructed with both machine and expert verification to maximize data availability and reliability. 
We use \dataset{} to investigate the role of clinical specialties in the knowledge-intensive scenario of medical QA. 
Our results show that
1) training on data from a clinical specialty \textit{does not necessarily lead to best performance on that specialty}, and
2) regardless of the specialty the LLM was fine-tuned on, token probabilities of clinically relevant terms \textit{increase consistently across all specialties}.
Thus, we hypothesize improvement gains, at least in our settings, are derived mostly from \textit{domain shifting} (e.g., general to medical) rather than specialty-specific knowledge injection, and suggest rethinking the role of fine-tuning data in the medical domain.
To motivate further advancements in the clinical NLP field, we release \dataset{} and all code needed to reproduce all our experiments to the research community.\footnote{\url{https://github.com/nlp4health-lab/S-MedQA}}
\end{abstract}

\input{parts/01-intro}
\input{parts/07-dataset-creation}

\input{parts/08-experiments}                                            
\input{parts/04-discussion}
\input{parts/06-related}

\input{parts/05-conclusions}

\input{parts/limitations}
\input{parts/acknowledgment}


\bibliography{custom}
\bibliographystyle{acl_natbib}

\appendix
\input{appendices/01-appendix}

\end{document}

%% file: parts/01-intro.tex
\section{Introduction}




Multiple-choice question-answering (QA) datasets 
are widely used to benchmark large language models (LLMs) in the medical domain~\citep{singhal2023large,labrak2024biomistral} and guide the development of medical LLMs (e.g., PubMedQA, \citealp[]{jin2019pubmedqa}; MASH-QA, \citealp[]{zhu-etal-2020-question}; MedQA, \citealp[]{jin2021disease}; MedMCQA, \citealp[]{pal2022medmcqa}). 
However, specialized hospitals may require LLMs to address specific clinical problems and are often interested in performance within \textit{one or a few clinical specialties} (e.g., obstetrics or oncology). 
To the best of our knowledge, no open-source medical QA datasets include clinical specialty annotations. 

\begin{figure*}[t!]
    \centering
    \includegraphics[width=\textwidth]{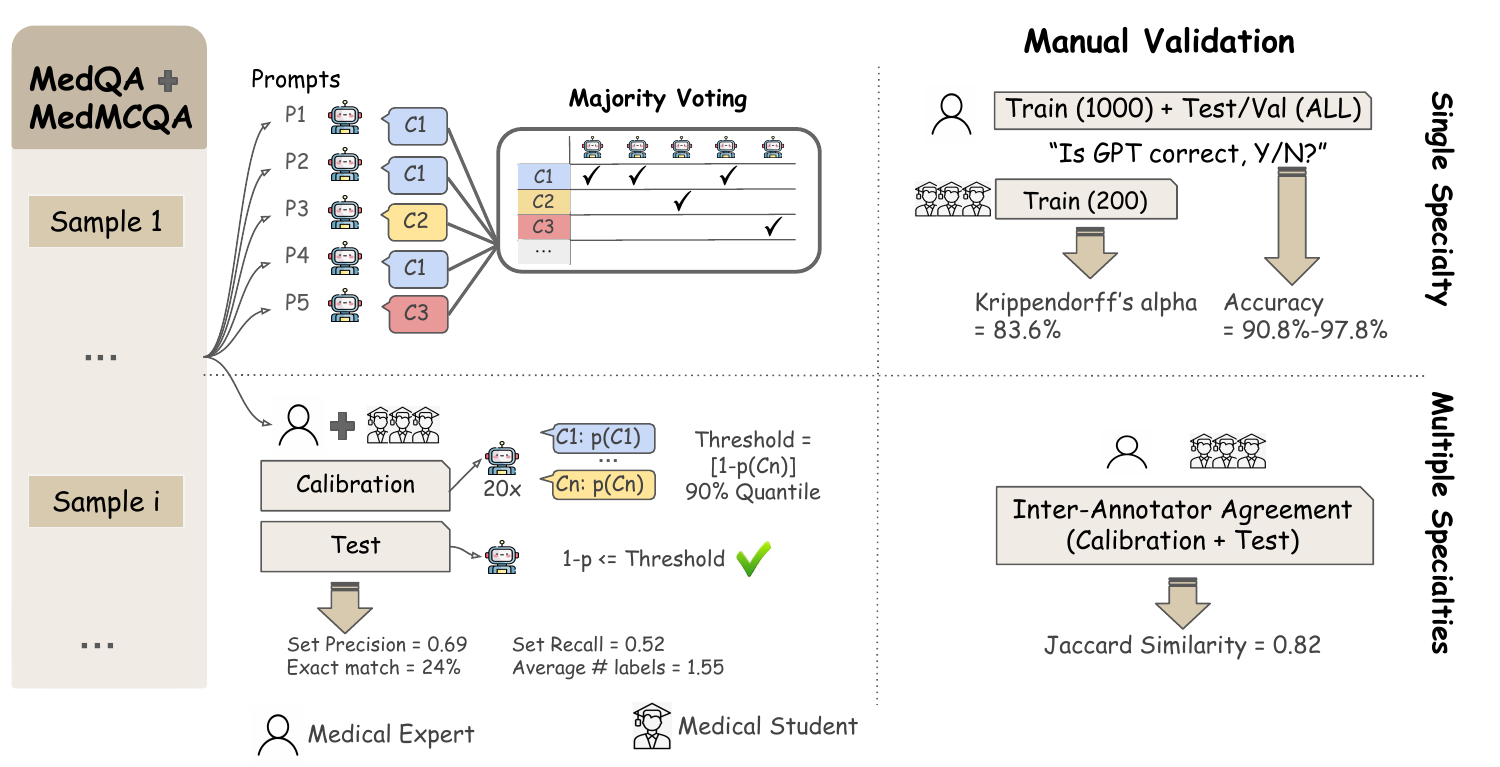}
    \caption{Overview of \dataset{}'s construction process. For single specialty annotation of each sample, we generate predictions using 5 different prompts and only keep those where predictions agree ($3+$, $4+$, or $5$ times).
    A medical expert manually annotates \dataset{}'s entire validation and test set.
    We randomly sample $1,000$ questions from our train set and ask multiple medical experts to evaluate \gptmodel{}'s predictions (we detail inter-annotator agreement in \S \ref{sec:manual_validation}), achieving accuracies ranging from $90.8$--$97.8$\%.
    For multi-specialty annotation, we leverage conformal prediction to assign labels across multiple clinical specialties, using a held-out set of $300$ samples manually annotated by the same medical expert and three medical students as calibration/test set, with a Jaccard Similarity of 0.82, reflecting a high inter-annotation agreement. It achieves a set precision of $0.69$, set recall of $0.52$, with $24\%$ exactly correct matches and an average prediction length of $1.55$.}

    \label{fig:overview}
\end{figure*}

To address this gap, we propose \textbf{\dataset{}}, the first English medical QA dataset with multiple clinical specialty annotations (see Figure~\ref{fig:overview} for an overview, and \S \ref{sec:dataset_creation} for details). We build \dataset{} based on the widely used MedQA~\citep{jin2021disease} and  MedMCQA~\citep{pal2022medmcqa} datasets, incorporating both machine and expert verification to semi-automatically map samples onto clinical specialties at scale, while guaranteeing the quality of the annotations.
We first use multiple prompts and a majority voting mechanism to label QA pairs with a single specialty.
\dataset{} includes 15 specialties, each with hundreds to thousands of samples, and single specialty expert validation shows up to 97.8\% accuracy in this annotation (see \S \ref{sec:dataset_creation} for details).
In order to account for cross-disciplinary questions, the second step is to expand single-specialty annotations into multiple specialties.
We use a multi-label conformal prediction procedure with coverage guarantees.
We annotate a held-out set of QA pairs with medical experts with a Jaccard Similarity of $0.82$, and build a conformal multi-label annotation procedure that leads to a precision of $0.69$, recall of $0.52$, $24$\% exactly correct predictions, and an average number of labels of $1.55$.


Different clinical specialties can have very different amounts of available data (e.g., see Figure~\ref{fig:distribution-all}).
We thus first use our dataset to address the following research question (\S \ref{sec:cross-specialty}): \textit{to what extent can LLMs use the knowledge learned from a clinical specialty to answer questions about other specialties?} 
We investigate~\citet{zhou2024lima}'s hypothesis that almost all knowledge in LLMs originates from pretraining, and that fine-tuning primarily serves to shift the model toward a specific knowledge domain.
To that regard, we first fine-tune LLMs on one clinical specialty and evaluate on all other specialties. 
Interestingly, we find that \textit{the best results often come from fine-tuning on unrelated clinical specialties}.
E.g., fine-tuning with \textit{neurology} data performs best on \textit{cardiology} domain, despite their knowledge 
being largely unrelated.
Moreover, although different pre-trained LLMs exhibit different knowledge transfer patterns across clinical specialties, neither of the LLMs we investigated show the best knowledge transfer when training and testing within a same clinical specialty. 

These results lead us to take a step back and ask: \textit{but are QA pairs from different clinical specialties really different?}
To answer this question, in \S \ref{sec:term-overlap-analysis} we curate the relevant clinical terms (e.g., diseases, procedures) for each clinical specialty using authoritative resources---such as the SNOMED-CT knowledge graph~\cite{Cornet2008-es}---to
effectively estimate the clinical knowledge overlap between QA pairs across clinical specialties.

Finally, we ask \textit{how does the probability of clinically-relevant terms for one specialty change before and after fine-tuning the LLM?}
We address this question in \S \ref{sec:token-probability}.
We 
again use the clinical terms curated in \S \ref{sec:term-overlap-analysis}
and analyze changes in term probabilities before and after fine-tuning on data from \textit{the same vs. different clinical specialties}.
Our results suggest that performance gains are driven more by domain shifts (from \textit{general} to \textit{medical}) than by fine-grained clinical knowledge injection.



Our main contributions are:
\begin{itemize}
    \item We introduce \dataset{}, a medical QA dataset spanning 15 clinical specialties with $24$k QA pairs annotated with high quality and human validated specialty annotations.
    \item We consider both single-specialty annotations and multiple-specialty labels to account for cross-disciplinary QA pairs.
    \item We systematically evaluate the impact of fine-tuning 8 LLMs on cross-specialty medical QA performance. Our findings indicate that performance gains in medical QA tasks are primarily driven by domain shifts rather than the injection of specialty-specific knowledge.
    \item We release S-MedQA and all code necessary for reproduction, providing the community with a valuable resource for benchmarking and improving medical LLMs.
\end{itemize}




\begin{figure*}[t!]
\centering
\includegraphics[width=\textwidth]{figures/specialty_counts.pdf}
\caption{The distribution of all specialties classified by GPT-3.5. The dark blue specialties are the 15 we finally included in our benchmark. Note that we conduct experiments on the 6 common specialties for simplicity.}
\label{fig:distribution-all}
\end{figure*}

%% file: parts/07-dataset-creation.tex
\section{Creation of \dataset{}}
\label{sec:dataset_creation}

We now describe the creation of \dataset{}, a high-quality benchmark for medical QA with clinical specialty annotations.
In Figure~\ref{fig:overview}, we show an overview of \dataset{}'s creation process.

Creating \dataset{} involves 1) using an LLM to produce single medical specialty annotations for each QA pair (\S \ref{sec:single_specialty_classification}),
2) a thorough human validation step to ascertain the quality and correctness of the annotations (\S \ref{sec:manual_validation}), and finally
3) using an LLM to extend these single specialty annotations to possibly multiple specialties whenever the question requires so drawing on ideas from conformal prediction (\S \ref{sec:multiple_specialty_classification}).
We release multiple versions of \dataset{} with varying accuracy/coverage trade-offs, controlled by majority voting thresholds  for including examples.
Users can opt for a \textit{cleaner dataset with fewer samples} or a \textit{noisier one with more samples}.

\paragraph{Data and splits.}
We source examples from MedQA~\citep{jin2021disease} and MedMCQA~\citep{pal2022medmcqa}, two widely used medical QA datasets. MedQA samples follow their original train/valid/test splits, whereas we only use MedMCQA's training split as its test labels are not public.






\subsection{Clinical specialty categorization}
\label{sec:single_specialty_classification}
We consider the 55 medical specialties recognized in the European Union for labeling (see \S \ref{appendix-EU-specialty}).

Manually labeling QA examples with clinical specialties is costly and time-consuming.
We thus first use \gptmodel{} to annotate samples with single clinical specialties.
Preliminary experiments using a single prompt to predict a single specialty produced low accuracy annotations ($\sim75$\%). 
To improve this, we design five prompts (details in \S \ref{appendix_gpt_prompts}) and predict single specialties with \gptmodel{} for each QA pair independently with each prompt.
We then apply majority voting to decide on the specialty for that QA pair~\citep{ding2022gpt,goel2023llms}.
Human evaluation shows accuracies between $90.8$\%--$97.8$\% (more details next in \S \ref{sec:manual_validation}).





We exclude
$3,518$ ($11.5$\%) samples categorized as \textit{Others} since they mostly contain irrelevant clinical information (see Appendix~\ref{appendix:example-exclude} for examples of exclusion),
and then focus on $15$ out of $55$ specialties with more than $500$ samples.
The final dataset comprises $24306$ / $899$ / $899$ samples in train/validation/test sets after human validation.

In Figure~\ref{fig:distribution-all}, we show the distribution of samples across specialties. We show the 15 specialties we include in \dataset{} in \textcolor{Blue}{dark blue}, comprising in total 70.0\% / 70.7\% / 70.1\% of the entire \mbox{train / valid / test} sets.
In Table~\ref{tab:dataset-description}, we show the 15 specialties included in \dataset{} with the respective numbers of samples.
The columns represents majority voting with 3+, 4+, and 5 prompts.

\begin{table}[t!]
\centering
\resizebox{\linewidth}{!}{
\begin{tabular}{>{\raggedleft\arraybackslash}m{11em} rrr}
\toprule
Number of Votes (out of 5) & 3+ & 4+ & 5 \\
\midrule
Cardiology                   & 3,000 & 2,652 & 2,122 \\
Neurology                    & 2,954 & 2,419 & 1,627 \\
Endocrinology                & 2,384 & 2,122 & 1,568 \\
Infectious diseases          & 2,161 & 1,705 & 1,087 \\
Hematology                   & 2,005 & 1,746 & 1,332 \\
Gastroenterology             & 1,934 & 1,594 & 1,176 \\
Emergency medicine           & 1,597 & 1,098 &  629 \\
Pediatrics                   & 1,553 & 1,036 &  559 \\
Nephrology                   & 1,550 & 1,340 & 1,004 \\
Respiratory medicine         & 1,253 &  825 &  395 \\
Rheumatology                 & 1,115 &  958 &  700 \\
Obstetrics and gynecology    &  913 &  802 &  636 \\
Oncology                     &  770 &  570 &  300 \\
Psychiatry                   &  656 &  582 &  446 \\
Dermatology                  &  531 &  421 &  280 \\
\bf Total                        & \bf 24,306 & \bf 19,290 & \bf 14,261 \\
\bottomrule
\end{tabular}}
\caption{S-MedQA description. Number of samples of the 15 specialties using different minimum numbers of votes (3+, 4+, 5) in the train sets included in \dataset{}.}
\label{tab:dataset-description}
\end{table}

\subsection{Manual validation for single specialty predictions}
\label{sec:manual_validation}

\begin{table}[t!]
\centering
\scalebox{0.95}{
\begin{tabular}{@{}lccccc@{}}
\toprule
Prompts & \#1 & \#2 & \#3 & \#4 & \#5 \\
\midrule
Accuracy(\%)   & 76.0 & 72.8 & 73.0 &  73.8 & 80.2 \\
\bottomrule
\end{tabular}}
\caption{Accuracy of each single prompt. Prompts \#$1$ to \#$5$ are shown in Figures~\ref{fig:gpt_prompts_1}--\ref{fig:gpt_prompts_5} in Appendix~\ref{appendix_gpt_prompts}.}
\label{tab:classification-accuracy}
\end{table}

A medical expert labels all the examples in \dataset{}'s validation and test sets with the \textit{single most correct clinical specialty}. 
This expert also validates $1,000$ random samples from the train set, confirming whether the specialties predicted by \gptmodel{} are correct.\footnote{This is the specialty that is the most relevant to the QA pair. In this evaluation, we exclude any cases where there is ambiguity in which clinical specialty is the most relevant.}
Table~\ref{tab:classification-accuracy} shows the categorization accuracies when using single prompts. The accuracies when applying majority voting are shown on top in Table~\ref{tab:accuracy-vs-coverage}.
In general, when using voting with multiple prompts, we see large performance gains compared to using single prompts. (e.g., from $72.8$--$80.2$\% to $90.8$--$97.8$\%).

\input{tables/majority-voting-accuracy-vs-coverage}

In Table~\ref{tab:accuracy-vs-coverage}, we also show the accuracy vs. coverage trade-off over the $1,000$ random samples from the train set for different requirements for majority voting.
A higher quorum results in higher accuracy ($90.8 \rightarrow 97.8$) but greatly decreases the coverage
($89.1 \rightarrow 49.2$).
We release per-prompt single-specialty categorizations with votes for all examples for users to decide their preference between accuracy and coverage---more data but possibly more noise or less noise but less data---based on their specific use cases. We select `3+' as the quorum in this study for adequate fine-tuning data.

Moreover, to assess the trustworthiness of the medical expert, we randomly sample $200$ from the $1,000$ examples and further ask three medical graduate students to validate the same examples in the same procedure.
We use Krippendorff's alpha~\citep{hayes2007answering} to measure the inter-annotator agreement among the four annotators
and obtain 83.6\% (95\% CI [69.0\%, 93.9\%]).

\begin{figure*}[htbp]
    \centering
    \includegraphics[width=0.95\textwidth]{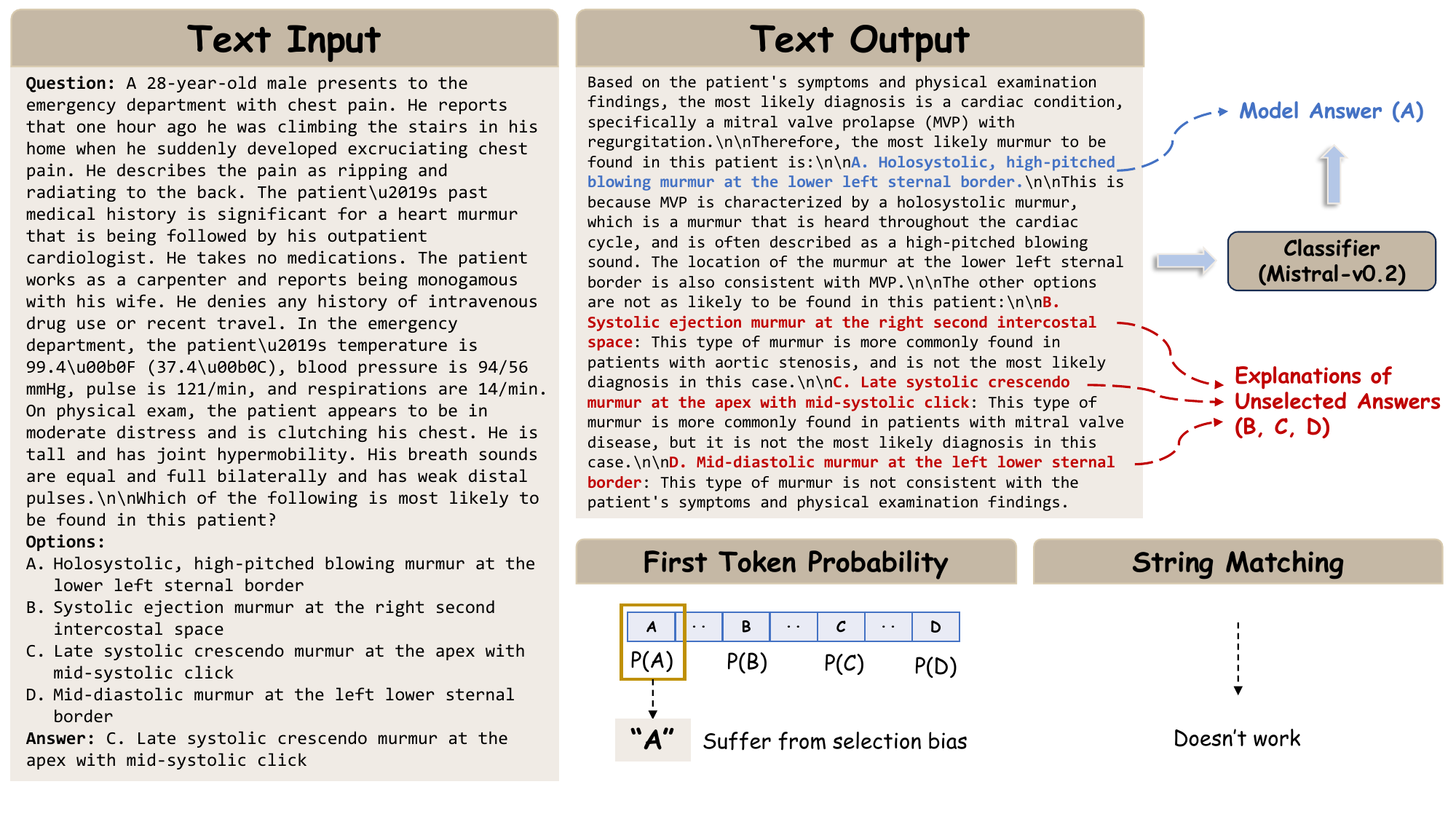}
    \caption{The illustration of first token probability, string matching, and our approach (classifier) to evaluating LLMs performance on S-MedQA. 
    We use text output instead of first token probability for evaluation because the latter suffers heavily from selection bias in multiple-choice QA~\citep{wang2024my}. However, string matching does not work in some cases. Our classifier trained on Mistral-v0.2 works successfully with an accuracy of 96.5\%.}
    \label{fig:classifier}
\end{figure*}

\subsection{From single to multiple clinical specialties}
\label{sec:multiple_specialty_classification}
We devise a strategy based on conformal prediction to expand clinical specialty annotations to possibly multiple specialties if a QA pair requires so.
Our main idea is: 1) \textit{calibrate} an LLM to find a threshold that maximises multiple specialties' prediction accuracy in a calibration set; 2) use this threshold to build a conformal classifier that predicts multiple clinical specialties for each QA pair.

\paragraph{Calibration and test set.}
We curate a held-out set of 300 samples, annotated independently by a medical expert (300 samples) and three medical students (100 samples each). We measure the Jaccard Similarity between the expert and the students for the inter-annotator agreement, with a similarity score of 0.82.
We use 200 examples (66.7\%) as a calibration set, and report results on the remaining 100 examples (33.3\%).
\paragraph{Conformal threshold and set construction.}

Following ~\citet{ke2025correctness}, we query GPT-3.5-Turbo $n=20$ times per input question and predict a single specialty $C_i$.
We then get frequencies $P(C_i)$ over all possible output specialties (see \S \ref{appendix-EU-specialty}).
We compute the non-conformity scores as $S_i=1-P(C_i)$ 
and obtain a set of scores $S_1$, $S_2$, ..., $S_n$ per question.
We set our acceptable error to 10\%  ($\alpha=0.1$) and then use the ($1-\alpha$) quantile of all the scores as the calibration threshold. 
At prediction time, we also sample 20 times single specialty categorization for each question, and compute the per specialty scores.
We include all specialties that have lower score than the calibration threshold in the final multi-specialty prediction. 

We report key metrics computed on the testing data.
The number of instances where the predicted label set exactly matched the true gold label set was $24$. Precision, defined as the proportion of predicted labels that were correct, was $0.69$. Recall, which measures the proportion of gold labels correctly included in the prediction sets, was $0.52$. The average number of predicted labels was $1.55$.

%% file: tables/majority-voting-accuracy-vs-coverage.tex
\begin{table}[t!]
\centering
\begin{tabular}{@{}lccc@{}}
\toprule
{Number of Votes (out of 5)} & $3+$ & $4+$ & $5$ \\
\midrule
Accuracy (\%)   & 90.8 & 94.8 & 97.8 \\
Coverage (\%)   & 89.1 & 69.0 & 49.2 \\
\bottomrule
\end{tabular}
\caption{Accuracy vs. coverage for majority voting under different minimum number of votes.}
\label{tab:accuracy-vs-coverage}
\end{table}

%% file: parts/08-experiments.tex
\section{Experiments}
\label{sec:experiments}

We first investigate how training LLMs on data from a clinical specialty impacts their performance on other specialties (\S \ref{sec:cross-specialty}). Our findings show that training on one clinical specialty does not necessarily lead to the best performance in that specialty. We hypothesize whether QA pairs from different clinical specialties are indeed very different from each other. That leads us to quantify the overlap of clinically relevant terms in train/test splits \textit{within} and \textit{across} specialties (\S \ref{sec:term-overlap-analysis}).
Finally, we investigate to what extent token probabilities of clinically relevant tokens change when fine-tuning LLMs in \dataset{} compared to non-clinical data (\S \ref{sec:token-probability}).




\input{tables/results-llama3.1}

\subsection{Experimental setup}
To minimize potential evaluation bias, we follow best practices to evaluate LLM performance on QA datasets (1) by shuffling the order of the answers multiple times and adding multiple shuffled QA pairs in the test set, and (2) by using the entire answer to a question instead of the LLM's maximum token probability among options A, B, C, D in the answer~\citep{zheng2023large,wang2024my}.

More specifically, in all test sets, we shuffle the answers 5 times for each sample and add all these 5 entries to the final test set in case the model prefers an option due to its position~\citep{zheng2023large}. 
To further improve the reliability, we follow~\citet{wang2024my} to train a classifier to match model outputs to the options in a post-hoc step, instead of using the maximum probability of options \{A, B, C, D\} with a single next-token prediction step. More concretely, we randomly select 150 training samples and generate answers for these with four LLMs ({Llama2-7b}, {Llama2-13b}, {Mistral-v0.1}, and {Mistral-v0.2}), resulting in $600$ responses.
We manually annotate all the responses with the right options and use these annotations to train a Mistral-Instruct-v0.2 model as the classifier,
with $400$ ($200$) train (test) samples. Our classifier achieves 96.5\% accuracy and we use it in all experiments.

Figure~\ref{fig:classifier} illustrates the approach (classifier) we use to evaluate the performance of the models and addresses the issues with using the first token probability or simple string matching~\citep{wang2024my}. The classifier is trained based on \textbf{Mistral-v0.2} and applied in all experiments.

In all subsequent analyses below, we use the 6 most common specialties (\textit{Cardiology}, \textit{Gastroenterology}, \textit{Infectious diseases}, \textit{Neurology}, \textit{Obstetrics and Gynecology}, and \textit{Pediatrics}). The number of test samples in each specialty is $80$, $83$, $102$, $74$, $88$, and $90$, respectively. The final number of samples in the test set is $400$, $415$, $510$, $370$, $440$, and $450$ (i.e., times 5 shuffles).

\subsection{Cross-specialty evaluation}
\label{sec:cross-specialty}

We experiment with 8 open-source LLMs: \textbf{Llama2-Chat-7B}, \textbf{Llama2-Chat-13B}, \textbf{Llama-3.1-8B-Instruct}, \textbf{Llama-3.2-3B-Instruct}~\citep{touvron2023llama}, \textbf{Mistral-Instruct-v0.1}, \textbf{Mistral-Instruct-v0.2}~\citep{jiang2023mistral} , \textbf{Bio-Medical-LLaMA-3-8B}~\citep{ContactDoctor_Bio-Medical-Llama-3-8B}, and \textbf{OLMo-2-1124-7B-Instruct}~\citep{olmo20252olmo2furious}.
These include six general-purpose models (Llama-2, Llama-3, and Mistral-based models), one biomedical Llama model, and a fully open-source model (OLMo).

We fine-tune each LLM on the six per-specialty training sets using prompts from \S \ref{appendix-llm-prompt}, and evaluate on all six test sets. Models are trained for up to 10 epochs, with selection based on per-specialty validation accuracy. We also train on the combined dataset to evaluate how exposure to larger and more diverse data affects models' performance.
We use LoRA~\citep{hu2021lora} on all projection layers for the fine-tuning process in all experiments with the following hyperparameters: learning rate $2$e-$5$, rank $32$, alpha $16$, dropout rate $0.1$, batch size $8$.

\paragraph{Results.}

Table~\ref{tab:6x6-llama3.1} shows the performance of Llama-3.1-8B-Instruct fine-tuned independently on each specialty and tested on all six specialty test sets, as well as after fine-tuning on a combination of all six specialties.
\textbf{Are improvements truly indicative of knowledge acquisition or injection?}
Notably, for models fine-tuned on individual specialties,
\textit{none of the best performing models were trained on the corresponding specialty's training data}, e.g., the best performance on the \textit{Cardio} test set (69.9\%) was achieved by the model trained on \textit{Neuro}.
If the improvements were due to knowledge injection, we would expect to see best-performing models consistently along the diagonal (e.g., \textit{Cardio} $\rightarrow$ \textit{Cardio}, \textit{Infect} $\rightarrow$ \textit{Infect}, and so on). 
We report results for other LLMs in \S \ref{appendix_result_three} and note that similar inconsistencies hold, whereas with different transfer patterns across specialties. 
This pattern suggests that performance gains from fine-tuning primarily arise from factors other than knowledge injection, partially supporting \citet{zhou2024lima}'s hypothesis in the clinical domain. 

The number of per-specialty examples used to train Llama-3.1-8B-Instruct is imbalanced (e.g., see Table~\ref{tab:dataset-description}). 
To tackle this imbalance, we randomly select 913 samples from each of the six specialties' training data, which corresponds to the least number of samples among the six specialties. We fine-tune Llama-3.1-8B-Instruct on this data and we show the results in \S\ref{appendix-robustness}.
We also conduct 20 bootstrap resampling experiments on the test sets using the Llama-3.1-8B-Instruct model fine-tuned on the whole (i.e., imbalanced) training data to tackle the issue of limited testing data. We randomly sample the same number of samples as the test sizes from the original test sets with replacement. We show the results for this experiment in \S\ref{appendix-robustness}.
Both experiments show that most of the best performing models on each per-specialty's test set is not the same model tuned on the training data from the same specialty.
This demonstrates the robustness of our main results, i.e., best models are almost always off-diagonal.

\subsection{Term overlap analysis}
\label{sec:term-overlap-analysis}
Clinical questions from different specialties can differ substantially in terms of knowledge; for example, one cannot assume that \textit{neurology} expertise can directly apply to \textit{cardiology}. Here, we thus quantify the overlap of clinically relevant terms in train/test splits \textit{within} and \textit{across} specialties. 
We do this by extracting clinically relevant terms in each question of \dataset{}, by mapping these terms onto relevant clinical specialties within the scope of the top 6 specialties we selected for further experiments, and finally by quantifying the difference in clinical terminology across different specialties. 

\begin{figure}[t]
    \centering
    \includegraphics[width=0.48\textwidth]{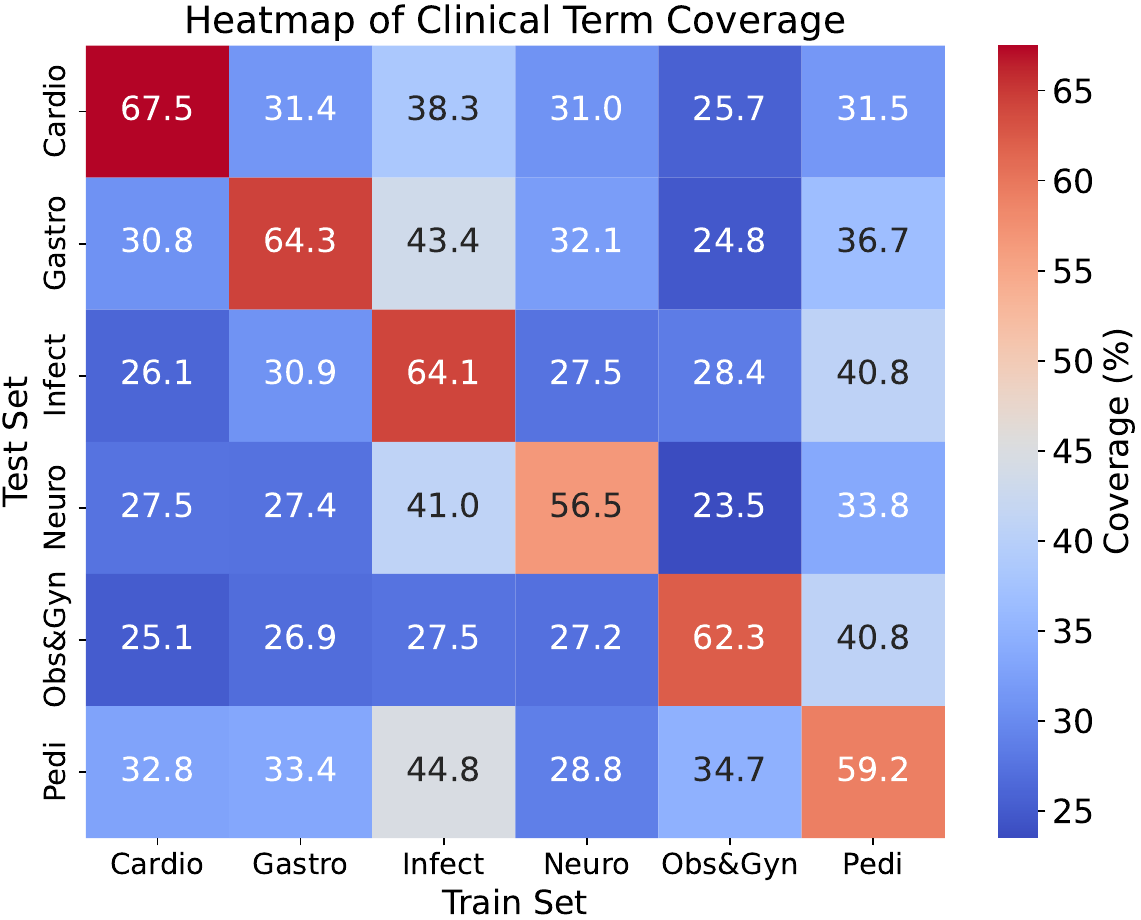}
    \caption{Heatmap showing the overlap of clinical terms within and across clinical specialties between train and test QA pairs in \dataset{}.}
    \label{fig:term-overlap-heatmap}
\end{figure}

We map each term in a question to relevant clinical specialties using SNOMED-CT~\citep{Cornet2008-es}\footnote{\url{https://www.snomed.org/}} and optionally the Human Phenotype Ontology~\citep[HPO;][]{castellanos2024human}.
1) We first manually identify the top-level \textit{disorder} concepts in SNOMED-CT that correspond to each medical specialty, and use the SNOMED hierarchy to link specific diseases or concepts to this top-level category.
E.g., we link any disorder under \textit{Disorder of cardiovascular system} as part of \textit{cardiology}.\footnote{For all high-level SNOMED-CT concepts we use for each clinical specialty, please refer to \S~\ref{appendix-snomed-concepts}.}
We found no simple mechanism to map \textit{findings}, \textit{procedures}, and \textit{observable entities} concepts to disorders in SNOMED-CT directly.
We thus first search \textit{findings}, \textit{procedures}, and \textit{observable entities} in HPO, identify the related disorders, which we then use to map onto clinical specialties similarly with SNOMED-CT.
2) For each question, we then use SciSpacy~\cite{neumann-etal-2019-scispacy} to perform entity linking of \textit{disorder}, \textit{finding}, \textit{procedure}, and \textit{observable entity} mentions in the question to SNOMED-CT.
3) Finally, we use the top-level concepts identified in (1) and the SNOMED-CT hierarchy to link each mention to a clinical specialty.

We exclude from this analysis general terms unrelated to specific specialties or shared by four or more of the six specialties.

\paragraph{Results.}
Figure~\ref{fig:term-overlap-heatmap} illustrates the overlap percentages of specialty-specific terminology between training and test sets for the six specialties. The average overlap in train/test splits within the same specialty is 63.4\%, indicating that training sets cover well clinical terminologies necessary
for solving corresponding test set questions. In contrast, the average overlap across different specialties is only 32.8\%, showing the domain-specific nature of medical terminology and the limited commonality between specialties. 
These results confirm that  ``knowledge leakage'' during cross-specialty evaluations is minimal, as different specialties share limited common specialty-specific terminology.

\subsection{Token probability}
\label{sec:token-probability}

We hypothesize that the performance improvements in Table~\ref{tab:6x6-llama3.1} are primarily due to domain shifting from general LLMs to the clinical domain, rather than the injection of new clinical knowledge.
We demonstrate this point through a detailed analysis of token probability shifts for clinical terms before and after fine-tuning.

\paragraph{The impact of medical data fine-tuning.}

To assess that, we analyze token probability changes for clinical terms in the questions from test sets linked to a single clinical specialty between a baseline model and the same model fine-tuned on data from each different specialty. Clinically relevant terms are extracted and mapped to specialties using the method described in \S\ref{sec:term-overlap-analysis}, and each term’s probability is obtained by summing the probabilities of its constituent subword tokens.

\begin{figure}[t!]
    \centering
    \includegraphics[width=0.48\textwidth]{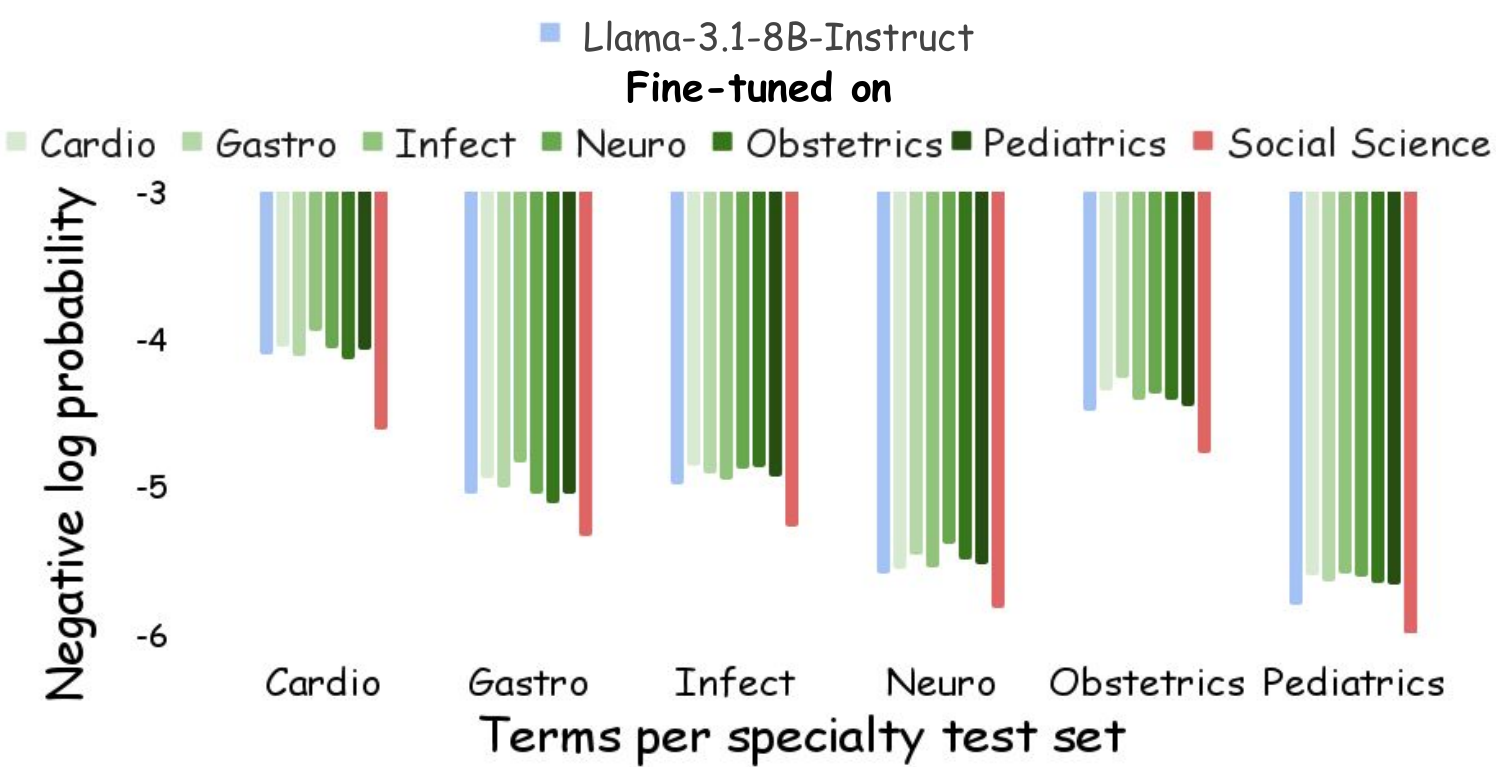}
    \caption{Negative log-probabilities for clinically relevant tokens between baseline Llama-3.1-8B-Instruct and the same model further fine-tuned on each specialty data. Each group represents tokens categorized into different clinical specialties. Each color means that the same model is further fine-tuned on each specialty data.}
    \label{fig:token-probability}
\end{figure}

In Fig.~\ref{fig:token-probability} we show the average log-probabilities of medical terms across specialties as predicted by Llama-3.1-8B-Instruct (base and fine-tuned) for each of the six specialties.
Regardless of the specialty used for fine-tuning, we observe similarly increased token log-probabilities for clinical terms specific to the fine-tuning specialty, as well as for terms associated with other specialties.

We note that token log-probabilities for terms from different specialties differ in range, likely reflecting the pre-trained model's existing knowledge distribution. This also seems to support our hypothesis that fine-tuning shifts the domain rather than injects new domain knowledge.


\paragraph{Are improvements due to additional training steps?}



To rule out the possibility that the similarly increased probabilities in all clinical specialties are due simply to the additional training steps, we also fine-tune our baseline Llama-3.1-8B-Instruct on three social sciences subsets of MMLU~\citep{hendryckstest2021,hendrycks2021ethics}---public relations, security studies, and sociology---which are entirely unrelated to the medical domain.
Again, we compare the average token log-probabilities of the six specialty medical terms produced by this model with those obtained from models fine-tuned on the corresponding S-MedQA datasets.
When fine-tuning on MMLU out-of-domain datasets there is a significant drop in token probabilities compared to models trained on S-MedQA data (red bars in Figure~\ref{fig:token-probability}).
We thus confirm that this drop is caused by domain shifting rather than purely the effect of additional training.
See Appendix~\ref{appendix:percentage-token-probability} for details of the token probabilities for models trained on \dataset{} and MMLU social sciences.

%% file: tables/results-llama3.1.tex
\begin{table*}[t]
\centering
\scalebox{0.8}{
\begin{tabular}{>{\centering\arraybackslash}m{1em} >{\raggedright\arraybackslash}m{7em} >{\centering\arraybackslash}m{4em} >{\centering\arraybackslash}m{4em} >{\centering\arraybackslash}m{4em} >{\centering\arraybackslash}m{4em} >{\centering\arraybackslash}m{4em} >{\centering\arraybackslash}m{4em} >{\centering\arraybackslash}m{4em}}

& \textbf{Test Sets} & Cardio & Gastro & Infect & Neuro & Obstetrics & Pediatrics & avg. \\
\cmidrule{2-9}
\multirow{8}{*}{\STAB{\rotatebox[origin=c]{90}{Train Sets}}} & Llama-3.1$^\dagger$ & 64.0 & 67.5 & 65.3 & 66.8 & 65.0 & 64.5 & 65.5 \\
\cmidrule{3-9}
& Cardio & \underline{69.3} & \textbf{76.5} & \textbf{73.5} & 76.2 & \textbf{}{70.5} & 70.9 & 72.8 \\
& Gastro & 68.2 & \underline{73.3} & 71.8 & 74.6 & 69.3 & 69.2 & 71.1 \\
& Infect & 65.8 & 74.2 & \underline{69.5} & 72.5 & 66.6 & 66.7 & 69.3 \\
& Neuro & \textbf{69.9} & 76.4 & 72.3 & \underline{\textbf{76.9}} & 69.5 & \textbf{71.7} & 72.8 \\
& Obstetrics & 69.8 & 75.0 & 72.9 & 75.4 & \underline{68.7} & 69.4 & 71.9 \\
& Pediatrics & 66.9 & 75.0 & 72.1 & 75.9 & 68.7 & \underline{70.2} & 71.4 \\
\cmidrule{3-9}
& Combined$^\ddagger$ & 72.9 & 79.5 & 75.4 & 77.8 & 70.8 & 74.1 & 75.1 \\
\cmidrule{2-9}
\end{tabular}}
\caption{Accuracy matrix for Llama-3.1-8B-Instruct. 
$^\dagger$Model is applied without finetuning. 
$^\ddagger$Model is trained on the combination of all 6 specialty train sets. 
For robustness, accuracies are computed as the average over the test set with answer options shuffled five times.
For each specialty, the best performance when fine-tuned on different specialty datasets is in \textbf{bold}, and scores for models fine-tuned on the same specialty are \underline{underlined}. Surprisingly, none of the best performances come from models fine-tuned on their corresponding training sets.}
\label{tab:6x6-llama3.1}
\end{table*}

%% file: parts/04-discussion.tex
\section{Discussion}
\label{sec:discussion}
The effectiveness of lightweight fine-tuning on strong base models has been widely recognized in NLP, largely following the pretraining paradigm introduced in the BERT era~\citep{devlin-etal-2019-bert}. This strength is often attributed to domain shifting, especially for linguistically focused tasks such as sentiment analysis~\citep{zhang2020semantics}, natural language inference~\citep{koroteev2021bert}, and semantic matching~\citep{gao2021simcse}.

In knowledge-intensive domains such as healthcare, the role of fine-tuning remains underexplored. Clinical questions across specialties often pose substantial knowledge barriers, and the capabilities of current LLMs on specialty-specific tasks are still unclear. Moreover, it remains uncertain whether these tasks benefit more from explicit knowledge injection or from effective knowledge transfer within the base model. This paper benchmarks these real-world questions in a fine-grained manner and provides preliminary evidence to explain the role of fine-tuning in the clinical domain.

Moreover, current LLMs are trained on massive datasets and raise concerns about benchmark contamination. Our experiments span a wide range of models, including open-source ones such as OLMo~\citep{groeneveld-etal-2024-olmo}, which is trained without clinical data. Similar findings on these models strengthen the reliability of our results.


%% file: parts/06-related.tex
\section{Related Work}
\label{sec:related}

\paragraph{Medical QA datasets.}
Medical QA datasets have played a pivotal role in benchmarking and advancing the capabilities of LLMs in the healthcare domain. Early datasets such as PubMedQA~\citep{jin2019pubmedqa} and MedQA~\citep{jin2021disease} introduced multiple-choice QA tasks derived from medical board exams, focusing on assessing clinical reasoning and factual knowledge. MedMCQA~\citep{pal2022medmcqa} later expanded the scope with a larger dataset covering a wide range of medical subjects. These datasets have been instrumental in developing models like BioGPT~\citep{luo2022biogpt} and MedPaLM~\citep{singhal2025medpalm}, which leverage domain-specific pretraining to achieve human-like performance on medical tasks.

However, these datasets lack annotations for clinical specialties, limiting their applicability in 
scenarios where specialty-specific knowledge is crucial. Recent efforts, such as BioMistral~\citep{labrak2024biomistral}, have focused on pretraining biomedical LLMs on medical corpora but have not addressed the need for fine-grained specialty-level evaluation. S-MedQA fills this gap by introducing per-specialty annotations and enabling the study of knowledge transfer across specialties.

\paragraph{Knowledge injection vs. Domain shifting.}


The effectiveness of fine-tuning on task-specific datasets has been extensively studied in NLP~\citep{devlin-etal-2019-bert,mayfield-black-2020-fine,limsopatham-2021-effectively}. Broadly speaking, this effectiveness is often understood to either introduce new knowledge within the external corpus into pre-trained models~\citep{fu-etal-2023-revisiting,mecklenburg2024injecting}, especially in knowledge-intensive tasks, or to adapt a model’s general capabilities to more specialized domains~\citep{limsopatham-2021-effectively,garrido2023fine,wu-etal-2024-far}, highlighting effective knowledge sharing across tasks.

Particularly, recent studies~\citep{zhou2024lima} in LLMs suggest that fine-tuning primarily serves to shift the model's focus toward a specific domain, leveraging knowledge already encoded during pretraining, rather than injecting new knowledge into the model.
\citet{zhou2024lima} introduced the hypothesis that alignment tuning---e.g., supervised fine-tuning (SFT) and reinforcement learning from human feedback (RLHF)---primarily teaches large language models (LLMs) to select a sub-distribution of response styles and formats. This hypothesis was supported by their study on the LIMA model, where fine-tuning with as few as 1,000 examples yielded strong alignment. This study demonstrated that the success of alignment tuning lies in its ability to adjust stylistic tokens, transitional phrases, and disclaimers, while the majority of the knowledge required to answer user queries remains unchanged from the base model.
Our work is a first step towards investigating the role of fine-tuning in knowledge-intensive and finely segmented domains, since we specifically address the medical domain with different clinical specialties.




%% file: parts/05-conclusions.tex
\section{Conclusions}

In this paper, we introduce \dataset{}, the first English medical QA dataset annotated for 15 distinct clinical specialties including single- and multi-specialty annotations for cross-disciplinary questions.
We investigate different questions using \dataset{} and find that questions from different clinical specialties have significant differences in terms of clinical terminology.
Despite that, we also find that the best QA performance is almost always observed for LLMs fine-tuned on unrelated clinical specialties.
Moreover, we also show that fine-tuning the same base LLM on high-quality non-medical data leads to decreased token probabilities for clinically relevant terms.
Our results suggest that~\citet{zhou2024lima}'s hypothesis apply to the medical domain and that improvements can be primarily attributed to domain shifting rather than knowledge injection.
However, the precise impact of different types of QA data (e.g., complexity or difficulty of the QA pair) remains unclear.
Finally, we recommend further research to investigate the role of fine-tuning in the medical domain.


%% file: parts/limitations.tex
\section*{Limitations}
In the following, we discuss some limitations of our work.

\paragraph{Experiments on knowledge transfer patterns.}
Although we conduct experiments investigating how knowledge transfers within and across clinical specialties, we limited our experiments to the medical domain.
That means that validating to what extent \cite{zhou2024lima}'s hypothesis generalises to other knowledge-intensive domains is not something we address in this work.
Further research is needed to investigate the role of fine-tuning and instruction data in other domains.

\paragraph{Clinical specialties.}
We have included 15 clinical specialties in our dataset due to their availability in the source datasets we used.
This means that possibly relevant clinical specialties have been left out of our dataset (i.e., see the long tail in Figure~\ref{fig:distribution-all}).
Future work should aim to include QA pairs from more varied and possibly rarer clinical specialties (e.g., speech language pathology; ~\citealt{kim-etal-2024-medexqa}).

\paragraph{Language and healthcare system.}
Our dataset draws on examples in English extracted mostly from US-centric datasets.
We believe that important future work lies in extending QA datasets to other languages and healthcare systems.


%% file: parts/acknowledgment.tex
\section*{Acknowledgment}

XY and IC are funded by the project CaRe-NLP with file number NGF.1607.22.014 of the research programme AiNed Fellowship Grants which is (partly) financed by the Dutch Research Council (NWO).

%% file: appendices/01-appendix.tex
\section{Appendix}
\label{sec:appendix}

\subsection{Specialties recognized in the European Union (EU) and European Economic Area (EEA)}
\label{appendix-EU-specialty}
According to \textit{Directive 2005/36/EC of the European Parliament and of the Council of 7 September 2005 on the recognition of professional qualifications},\footnote{\url{https://eur-lex.europa.eu/legal-content/EN/ALL/?uri=CELEX\%3A32005L0036}} the following clinical specialties are recognized in the EU and EEA:
Allergist, Anaesthetics, Cardiology, Child psychiatry, Clinical biology, Clinical chemistry, Clinical microbiology, Clinical neurophysiology, Craniofacial surgery, Dermatology, Emergency medicine, Endocrinology, Family and General Medicine, Gastroenterologic surgery, Gastroenterology, General Practice, General surgery, Geriatrics, Hematology, Immunology, Infectious diseases, Internal medicine, Laboratory medicine, Nephrology, Neuropsychiatry, Neurology, Neurosurgery, Nuclear medicine, Obstetrics and gynecology, Occupational medicine, Oncology, Ophthalmology, Oral and maxillofacial surgery, Orthopedics, Otorhinolaryngology, Pediatric surgery, Pediatrics, Pathology, Pharmacology, Physical medicine and rehabilitation, Plastic surgery, Podiatric surgery, Preventive medicine, Psychiatry, Public health, Radiation Oncology, Radiology, Respiratory medicine, Rheumatology, Stomatology, Thoracic surgery, Tropical medicine, Urology, Vascular surgery, Venereology.

\subsection{Prompts used for specialty classification}\label{appendix_gpt_prompts}
In Figures~\ref{fig:gpt_prompts_1}--\ref{fig:gpt_prompts_5} we show the 5 prompts we use with \gptmodel{} for specialty classification. Prompt 1 is zero-shot, while we add 6 examples to the other prompts (one example from each top-6 specialty) to leverage the in-context ability of LLMs. We moved the list of specialties to the end of the user prompt in prompt 4 and changed the format of the user prompt to follow the examples by adding \textit{``Question:''} and \textit{``Answer:''} in prompt 5.

\subsection{Prompts used for LLM tuning and inferring}
\label{appendix-llm-prompt}


An example of the prompt we use for LLM tuning and inferring in all our experiments is as follows:\\
\newline [INST] Please read the multiple-choice question below carefully and select ONE of the listed options and only give a single letter.
\newline Question: A 62-year-old woman presents for a regular check-up. She complains of lightheadedness and palpitations which occur episodically. Past medical history is significant for a myocardial infarction 6 months ago and NYHA class II chronic heart failure. She also was diagnosed with grade I arterial hypertension 4 years ago. Current medications are aspirin 81 mg, atorvastatin 10 mg, enalapril 10 mg, and metoprolol 200 mg daily. Her vital signs are a blood pressure of 135/90 mm Hg, a heart rate of 125/min, a respiratory rate of 14/min, and a temperature of 36.5\u00b0C (97.7\u00b0F). Cardiopulmonary examination is significant for irregular heart rhythm and decreased S1 intensity. ECG is obtained and is shown in the picture (see image). Echocardiography shows a left ventricular ejection fraction of 39\%. Which of the following drugs is the best choice for rate control in this patient?
\newline A. Atenolol
\newline B. Diltiazem
\newline C. Propafenone
\newline D. Digoxin
\newline Answer: [/INST] D. Digoxin

\subsection{Excluded vs. complex examples}
\label{appendix:example-exclude}

\paragraph{Excluded examples}
We exclude examples classified as \textit{"Others"}, i.e., not belonging to any specialty in the given list of 55 specialties recognized by the EU. Here is an example:

\begin{quotation}
\textit{A resident in the department of obstetrics and gynecology is reading about a randomized clinical trial from the late 1990s that was conducted to compare breast cancer mortality risk, disease localization, and tumor size in women who were randomized to groups receiving either annual mammograms starting at age 40 or annual mammograms starting at age 50. One of the tables in the study compares the two experimental groups with regard to socioeconomic demographics (e.g., age, income), medical conditions at the time of recruitment, and family history of breast cancer. The purpose of this table is most likely to evaluate which of the following?}
\end{quotation}

This question belongs to \textit{Clinical Trial Design} instead of any listed clinical specialties and does not contain knowledge required for daily clinical practices. Similar cases also include \textit{Toxicology}, \textit{Epidemiology}, and \textit{Medical Ethics}. We thus exclude such samples from \dataset{}.

\paragraph{Complex examples}
We carefully look into the samples that did not reach a vote of three together with the medical expert and noticed that most of these examples are ambiguous in terms of medical specialties.
They are therefore difficult to be classified into one single specialty. For instance, many disagreements occur with \textit{Neurology} and \textit{Emergency Medicine} in an emergent neurological issue, such as the following question:

\begin{quotation}
\textit{A 78-year-old man is brought to the emergency department by ambulance 30 minutes after the sudden onset of speech difficulties and right-sided arm and leg weakness. Examination shows paralysis and hypoesthesia on the right side, positive Babinski sign on the right, and slurred speech. A CT scan of the head shows a hyperdensity in the left middle cerebral artery and no evidence of intracranial bleeding. The patient's symptoms improve rapidly after pharmacotherapy is initiated and his weakness completely resolves. Which of the following drugs was most likely administered?}
\end{quotation}

According to the expert, both \textit{Neurology} and \textit{Emergency Medicine} apply to this situation, as they contain clinical knowledge from both specialties and require collaboration of these two specialties in clinical practices. Also, classifying it exclusively into one of the specialties requires extra expertise that could be beyond the capabilities of \gptmodel{}, e.g. classify as \textit{Emergency Medicine} if the question itself mainly focuses on maintaining vital signs, and \textit{Neurology} when it comes to subsequent treatment phases. 
Such complex examples were the main reason why we decided to add multiple specialty annotations per question.

\subsection{Additional cross-specialty evaluation results}
\label{appendix_result_three}

In Tables~\ref{tab:6x6-llama3.1},~\ref{tab:6*6-llama7b},~\ref{tab:6*6-llama13b},~\ref{tab:6*6-mistral0.1},~\ref{tab:6x6-llama3.2},~\ref{tab:6x6-biomedllama3}, and~\ref{tab:6x6-olmo} we show cross-specialty evaluation matrices for Llama2-7b-chat, Llama2-13b-chat, Mistral-7b-instruct-v0.1, Llama-3.1-8B-Instruct, Llama-3.2-3B-Instruct, Bio-Medical-LLaMA-3-8B, and OLMo-2-1124-7B-Instruct in addition to our main results (in \S \ref{sec:cross-specialty}).
Here we also observe that in $\sim$70\% of the cases the best performance on each per-specialty test set is not achieved by the model that is tuned on training data from the same specialty.


\input{tables/results-llama7b}
\input{tables/results-llama13b}
\input{tables/results-mistralv0.1}
\input{tables/results-mistralv0.2}
\input{tables/results-llama3.2}
\input{tables/results-biollama}
\input{tables/results-olmo}

\input{appendices/prompt-1}
\input{appendices/prompt-2}
\input{appendices/prompt-3}
\input{appendices/prompt-4}
\input{appendices/prompt-5}

\subsection{Robustness of cross-specialty evaluation}
\label{appendix-robustness}

We carry out 2 experiments to show the robustness of our cross-specialty evaluation results in Table~\ref{tab:6x6-llama3.1}.
We show the results of our cross specialty evaluation fine-tuned on balanced training set sizes in Table~\ref{tab:equal_train}, and the results of the 20 bootstrap resampling experiments on the test sets in Table~\ref{tab:bootstrap}.
Both experiments show that most of the best performing models on each per-specialty's test set is not the same model tuned on the training data from the same specialty.
This demonstrates the robustness of our main results, i.e., best models are almost always off-diagonal.

\input{tables/equal_train}
\input{tables/bootstrap}

\subsection{High-level Concepts in SNOMED-CT}
\label{appendix-snomed-concepts}

The high-level SNOMED-CT \textit{disorder} concepts we use when mapping questions to clinical specialities are:
\textit{cardiology} (Disorder of cardiovascular system, \texttt{SCTID: 49601007}),
\textit{gastroenterology} (Disorder of digestive system, \texttt{SCTID: 53619000}),
\textit{infectious diseases} (Infectious disease, \texttt{SCTID: 40733004}),
\textit{obstetrics} (Disorder of female reproductive system, \texttt{SCTID: 363124003};
Disorder of fetus and/or mother during labor, \texttt{SCTID: 1269083008}),
\textit{neurology} (Disorder of nervous system, \texttt{SCTID: 118940003}),
\textit{pediatrics} (Behavioral and emotional disorder with onset in childhood, \texttt{SCTID: 231538003};
Disorder of fetus and/or newborn, \texttt{SCTID: 414025005};
Developmental disorder, \texttt{SCTID: 5294002}).

\subsection{Token probability change}
\label{appendix:percentage-token-probability}
Table~\ref{tab:specialty-decrease} shows the token negative log-likelihoods (the higher the better) for terms from each clinical specialty before and after fine-tuning on the six \dataset{} specialty data and MMLU social sciences data.
We note that, apart from a few exceptions, training on S-MedQA leads to consistently improved token probabilities for clinically relevant terms, whereas training on MMLU makes clinical relevant terms less likely.

\begin{table*}[t!]
\centering
\resizebox{0.7\linewidth}{!}{
\begin{tabular}{lcccccc}
\toprule
\multirow{2}{*}{\textbf{Models}} & \multicolumn{6}{c}{\textbf{Test Sets} ($\uparrow$ is better)} \\
& \textbf{Cardio}& \textbf{Gastro} & \textbf{Infect} & \textbf{Neuro} & \textbf{Obstetrics} & \textbf{Pediatrics} \\
\midrule
\textbf{Mistralv0.2-Base}       & -4.11           & -5.05           & -4.99           & -5.59           & -4.49              & -5.80              \\
\cmidrule{2-7}
\textbf{Cardio}                 & \green{-4.06}   & \green{-4.94}   & \green{-4.86}  & \green{-5.56}  & \green{-4.35}       & \green{-5.60}              \\ 
\textbf{Gastro}                 & \red{-4.12}     & \green{-5.01}   & \green{-4.91}  & \green{-5.46}  & \green{-4.27}       & \green{-5.65}              \\ 
\textbf{Infect}                 & \green{-3.95}   & \green{-4.84}   & \green{-4.96}  & \green{-5.55}  & \green{-4.41}       & \green{-5.59}              \\ 
\textbf{Neuro}                  & \green{-4.07}   & \green{-5.05}   & \green{-4.88}  & \green{-5.39}  & \green{-4.37}       & \green{-5.61}              \\
\textbf{Obstetrics}             & \red{-4.14}     & \red{-5.11}     & \green{-4.87}  & \green{-5.50}  & \green{-4.41}       & \green{-5.66}              \\ 
\textbf{Pediatrics}             & \green{-4.08}   & \green{-5.05}   & \green{-4.93}  & \green{-5.53}  & \green{-4.46}       & \green{-5.67}              \\
\cmidrule{2-7}
\textbf{MMLU}                   & \red{-4.62}     & \red{-5.34}     & \red{-5.27}    & \red{-5.83}     & \red{-4.78}         & \red{-6.01}              \\
\bottomrule
\end{tabular}}
\caption{Average token negative log-likelihood for terms from each clinical specialty before and after fine-tuning on the six \dataset{} specialty data and MMLU social sciences data. Best viewed in colour.}
\label{tab:specialty-decrease}
\end{table*}

%% file: tables/results-llama7b.tex
\begin{table*}[t]
\centering
\scalebox{0.75}{
\begin{tabular}{>{\centering\arraybackslash}m{1em} >{\raggedright\arraybackslash}m{7em} >{\centering\arraybackslash}m{4em} >{\centering\arraybackslash}m{4em} >{\centering\arraybackslash}m{4em} >{\centering\arraybackslash}m{4em} >{\centering\arraybackslash}m{4em} >{\centering\arraybackslash}m{4em} >{\centering\arraybackslash}m{4em}}

& \textbf{Test Sets} & Cardio & Gastro & Infect & Neuro & Obstetrics & Pediatrics & avg. \\
\cmidrule{2-9}
& Llama2-7b & 36.0 & 36.3 & 36.7 & 34.6 & 40.6 & 41.4 & 37.7 \\
\cmidrule{3-9}
\multirow{8}{*}{\STAB{\rotatebox[origin=c]{90}{Train Sets}}} & Cardio & \underline{34.3} & 29.1 & 32.6 & 28.3 & 31.5 & 29.5 & 31.0 \\
& Gastro & 31.3 & \textbf{\underline{32.5}} & 30.7 & 28.8 & \textbf{38.3} & 30.1 & 32.0 \\
&  Infect & \textbf{35.0} & 31.3 & \underline{32.8} & 31.0 & 33.3 & 29.0 & 32.1 \\
& Neuro & 32.8 & 26.3 & \textbf{35.4} & \textbf{\underline{31.3}} & 33.5 & \textbf{36.1} & 32.7 \\
& Obstetrics & 34.5 & 30.0 & 34.6 & 30.7 & \underline{37.9} & 33.2 & 33.6 \\
& Pediatrics & 30.5 & 27.8 & 34.1 & 25.8 & 33.5 & \underline{31.0} & 30.7 \\
\cmidrule{3-9}
& Combined & 42.0 & 40.1 & 38.5 & 37.2 & 42.5 & 36.1 & 39.4 \\
\cmidrule{2-9}
\end{tabular}}
\caption{Cross-specialty accuracy matrix of Llama2-7b.}
\label{tab:6*6-llama7b}
\end{table*}

%% file: tables/results-llama13b.tex
\begin{table*}[t]
\centering
\scalebox{0.75}{
\begin{tabular}{>{\centering\arraybackslash}m{1em} >{\raggedright\arraybackslash}m{7em} >{\centering\arraybackslash}m{4em} >{\centering\arraybackslash}m{4em} >{\centering\arraybackslash}m{4em} >{\centering\arraybackslash}m{4em} >{\centering\arraybackslash}m{4em} >{\centering\arraybackslash}m{4em} >{\centering\arraybackslash}m{4em}}
& \textbf{Test Sets} & Cardio & Gastro & Infect & Neuro & Obstetrics & Pediatrics & avg. \\
\cmidrule{2-9}
& Llama2-13b & 43.3 & 34.9 & 40.6 & 36.1 & 45.4 & 39.2 & 40.0 \\
\cmidrule{3-9}
\multirow{8}{*}{\STAB{\rotatebox[origin=c]{90}{Train Sets}}} & Cardio & \underline{38.3} & 34.7 & \textbf{41.1} & 28.5 & \textbf{38.8} & 31.8 & 35.8 \\
& Gastro & \textbf{38.5} & \textbf{\underline{35.8}} & 31.8 & 31.3 & 36.0 & 32.7 & 34.3 \\
& Infect & 36.0 & 31.3 & \underline{36.5} & 32.9 & 39.2 & 32.7 & 34.9 \\
& Neuroy & 31.3 & 29.3 & 36.2 & \underline{28.8} & 35.4 & 33.5 & 32.7 \\
& Obstetrics & 33.5 & 29.5 & 36.5 & 30.4 & \underline{36.0} & 32.7 & 33.3 \\
& Pediatrics & 37.5 & 34.9 & 37.0 & \textbf{33.2} & 38.5 & \textbf{\underline{36.1}} & 36.3 \\
\cmidrule{3-9}
& Combined & 44.0 & 45.9 & 42.4 & 40.2 & 45.8 & 42.9 & 43.6 \\
\cmidrule{2-9}

\end{tabular}}
\caption{Cross-specialty accuracy matrix of Llama2-13b.}
\label{tab:6*6-llama13b}
\end{table*}

%% file: tables/results-mistralv0.1.tex
\begin{table*}[t]
\centering
\scalebox{0.75}{
\begin{tabular}{>{\centering\arraybackslash}m{1em} >{\raggedright\arraybackslash}m{7em} >{\centering\arraybackslash}m{4em} >{\centering\arraybackslash}m{4em} >{\centering\arraybackslash}m{4em} >{\centering\arraybackslash}m{4em} >{\centering\arraybackslash}m{4em} >{\centering\arraybackslash}m{4em} >{\centering\arraybackslash}m{4em}}
& \textbf{Test Sets} & Cardio & Gastro & Infect & Neuro & Obstetrics & Pediatrics & avg. \\
\cmidrule{2-9}
& Mistral-v0.1 & 41.0 & 39.0 & 40.0 & 30.7 & 42.7 & 37.5 & 38.9 \\
\cmidrule{3-9}
\multirow{8}{*}{\STAB{\rotatebox[origin=c]{90}{Train Sets}}} & Cardio & \textbf{\underline{52.8}} & 46.1 & 47.7 & 39.9 & 47.9 & 44.3 & 46.6 \\
& Gastro & 48.3 & \textbf{\underline{50.4}} & 41.1 & 40.2 & 47.5 & 44.3 & 45.2 \\
& Infect & 51.5 & 42.7 & \textbf{\underline{49.0}} & \textbf{44.3} & 47.5 & 43.5 & 46.5 \\
& Neuro & 50.7 & 46.3 & 47.1 & \underline{44.0} & 49.6 & \textbf{48.9} & 47.8 \\
& Obstetrics & 45.8 & 44.2 & 44.3 & 41.3 & \textbf{\underline{53.8}} & 46.3 & 46.1 \\
& Pediatrics & 48.8 & 45.5 & 42.7 & 35.1 & 51.0 & \textbf{\underline{48.9}} & 45.5 \\
\cmidrule{3-9}
& Combined & 52.8 & 47.6 & 46.4 & 49.5 & 56.3 & 47.2 & 49.9 \\
\cmidrule{2-9}
\end{tabular}}
\caption{Cross-specialty accuracy matrix of Mistral-v0.1}
\label{tab:6*6-mistral0.1}
\end{table*}

%% file: tables/results-mistralv0.2.tex

\begin{table*}[t]
\centering
\scalebox{0.75}{
\begin{tabular}{>{\centering\arraybackslash}m{1em} >{\raggedright\arraybackslash}m{7em} >{\centering\arraybackslash}m{4em} >{\centering\arraybackslash}m{4em} >{\centering\arraybackslash}m{4em} >{\centering\arraybackslash}m{4em} >{\centering\arraybackslash}m{4em} >{\centering\arraybackslash}m{4em} >{\centering\arraybackslash}m{4em}}
& \textbf{Test Sets} & Cardio & Gastro & Infect & Neuro & Obstetrics & Pediatrics & avg. \\
\cmidrule{2-9}
& Mistral-v0.2 & 52.0 & 45.9 & 48.2 & 37.0 & 52.9 & 43.5 & 46.9 \\
\cmidrule{3-9}
\multirow{8}{*}{\STAB{\rotatebox[origin=c]{90}{Train Sets}}} & Cardio & \underline{51.8} & 54.6 & 44.3 & 47.6 & 51.5 & 44.6 & 49.4 \\
& Gastro & 54.9 & \underline{54.1} & 38.9 & 43.5 & 51.7 & 46.4 & 48.8 \\
& Infect & \textbf{55.4} & 53.9 & \underline{43.0} & \textbf{47.8} & 54.3 & 43.5 & 50.1 \\
& Neuro & 54.4 & \textbf{54.8} & 41.9 & \underline{45.7} & \textbf{56.0} & \textbf{49.0} & 50.8 \\
& Obstetrics & 54.4 & 52.4 & \textbf{45.7} & 41.9 & \underline{51.5} & 44.6 & 48.8 \\
& Pediatrics & 50.8 & 53.0 & 42.4 & 35.9 & 48.1 & \underline{46.1} & 46.5 \\
\cmidrule{3-9}
& Combined & 53.7 & 53.8 & 42.7 & 43.7 & 52.2 & 45.7 & 49.1 \\
\cmidrule{2-9}
\end{tabular}}
\caption{Cross-specialty accuracy matrix of Mistral-v0.1}
\label{tab:6*6-mistral0.2}
\end{table*}

%% file: tables/results-llama3.2.tex
\begin{table*}[t]
\centering
\scalebox{0.75}{
\begin{tabular}{>{\centering\arraybackslash}m{1em} >{\raggedright\arraybackslash}m{7em} >{\centering\arraybackslash}m{4em} >{\centering\arraybackslash}m{4em} >{\centering\arraybackslash}m{4em} >{\centering\arraybackslash}m{4em} >{\centering\arraybackslash}m{4em} >{\centering\arraybackslash}m{4em} >{\centering\arraybackslash}m{4em}}

& \textbf{Test Sets} & Cardio & Gastro & Infect & Neuro & Obstetrics & Pediatrics & avg. \\
\cmidrule{2-9}
\multirow{8}{*}{\STAB{\rotatebox[origin=c]{90}{Train Sets}}} & Llama-3.2 & 54.0 & 60.0 & 58.5 & 61.0 & 57.0 & 55.5 & 57.7 \\
\cmidrule{3-9}
& Cardio & \underline{59.8} & \textbf{68.2} & \textbf{67.9} & \textbf{70.6} & \textbf{62.5} & \textbf{59.0} & 64.7 \\
& Gastro & 58.0 & \underline{65.3} & 66.2 & 68.7 & 60.2 & 56.5 & 62.5 \\
& Infect & 59.5 & 63.9 & \underline{65.7} & 69.0 & \textbf{62.5} & 57.8 & 63.1 \\
& Neuro & 59.8 & 64.6 & 66.9 & \underline{69.8} & 62.2 & 58.6 & 63.6 \\
& Obstetrics & 60.1 & 65.6 & 67.1 & 70.2 & \underline{61.6} & 58.9 & 63.9 \\
& Pediatrics & \textbf{61.0} & 66.3 & 67.7 & 70.3 & 62.0 & \underline{58.6} & 64.3 \\
\cmidrule{3-9}
& Combined & 65.4 & 69.9 & 71.3 & 75.5 & 66.9 & 65.0 & 68.9 \\
\cmidrule{2-9}
\end{tabular}}
\caption{Cross-specialty accuracy matrix of Llama-3.2-3B-Instruct.}
\label{tab:6x6-llama3.2}
\end{table*}

%% file: tables/results-biollama.tex
\begin{table*}[t]
\centering
\scalebox{0.75}{
\begin{tabular}{>{\centering\arraybackslash}m{1em} >{\raggedright\arraybackslash}m{7em} >{\centering\arraybackslash}m{4em} >{\centering\arraybackslash}m{4em} >{\centering\arraybackslash}m{4em} >{\centering\arraybackslash}m{4em} >{\centering\arraybackslash}m{4em} >{\centering\arraybackslash}m{4em} >{\centering\arraybackslash}m{4em}}

& \textbf{Test Sets} & Cardio & Gastro & Infect & Neuro & Obstetrics & Pediatrics & avg. \\
\cmidrule{2-9}
\multirow{8}{*}{\STAB{\rotatebox[origin=c]{90}{Train Sets}}} & Bio-Llama & 70.1 & 72.3 & 71.0 & 72.8 & 70.2 & 69.4 & 71.0 \\
\cmidrule{3-9}
& Cardio & \underline{\textbf{79.7}} & 82.7 & 80.3 & 82.8 & 76.4 & \textbf{75.3} & 79.6 \\
& Gastro & 77.7 & \underline{\textbf{83.1}} & \textbf{81.5} & 82.8 & 76.6 & 74.9 & 79.5 \\
& Infect & 76.8 & 82.2 & \underline{78.9} & 82.2 & 75.1 & 72.8 & 78.1 \\
& Neuro & 77.1 & 80.7 & 79.7 & \underline{\textbf{83.8}} & 76.3 & 74.0 & 78.6 \\
& Obstetrics & 78.0 & 81.7 & 81.2 & 81.8 & \underline{76.6} & 73.7 & 78.9 \\
& Pediatrics & 79.3 & 82.3 & 81.3 & 82.8 & \textbf{78.2} & \underline{75.3} & 79.9 \\
\cmidrule{3-9}
& Combined & 82.7 & 84.3 & 82.0 & 85.5 & 79.6 & 79.5 & 82.3 \\
\cmidrule{2-9}
\end{tabular}}
\caption{Cross-specialty accuracy matrix of Bio-Medical-Llama-3-8B.}
\label{tab:6x6-biomedllama3}
\end{table*}

%% file: tables/results-olmo.tex
\begin{table*}[t]
\centering
\scalebox{0.75}{
\begin{tabular}{>{\centering\arraybackslash}m{1em} >{\raggedright\arraybackslash}m{7em} >{\centering\arraybackslash}m{4em} >{\centering\arraybackslash}m{4em} >{\centering\arraybackslash}m{4em} >{\centering\arraybackslash}m{4em} >{\centering\arraybackslash}m{4em} >{\centering\arraybackslash}m{4em} >{\centering\arraybackslash}m{4em}}

& \textbf{Test Sets} & Cardio & Gastro & Infect & Neuro & Obstetrics & Pediatrics & avg. \\
\cmidrule{2-9}
\multirow{8}{*}{\STAB{\rotatebox[origin=c]{90}{Train Sets}}} & OLMo & 36.0 & 39.5 & 40.3 & 38.8 & 37.2 & 35.9 & 38.0 \\
\cmidrule{3-9}
& Cardio & \underline{40.7} & 46.0 & 47.2 & 44.4 & \textbf{41.6} & 38.2 & 43.1 \\
& Gastro & 39.6 & \underline{46.4} & 47.7 & \textbf{45.7} & 41.0 & 35.7 & 42.8 \\
& Infect & 38.4 & 45.5 & \underline{\textbf{49.0}} & 45.7 & 40.2 & 37.2 & 42.7 \\
& Neuro & 39.0 & 46.4 & 46.5 & \underline{45.4} & 40.7 & \textbf{38.4} & 42.8 \\
& Obstetrics & \textbf{41.5} & 44.5 & 46.1 & 45.3 & \underline{39.5} & 37.6 & 42.4 \\
& Pediatrics & 39.6 & \textbf{46.9} & 48.5 & 43.5 & 39.7 & \underline{37.8} & 42.7 \\
\cmidrule{3-9}
& Combined & 48.7 & 54.1 & 55.6 & 55.5 & 48.1 & 42.2 & 50.8 \\
\cmidrule{2-9}
\end{tabular}}
\caption{Cross-specialty accuracy matrix of OLMo-2-1124-7B-Instruct.}
\label{tab:6x6-olmo}
\end{table*}

%% file: appendices/prompt-1.tex
\begin{figure*}
\begin{tcolorbox}
\caption{Prompt-1}
\label{fig:gpt_prompts_1}
\vspace{0.3cm}
\begin{center}
\begin{minipage}{0.95\textwidth}
\scriptsize
\textbf{\#\#\# System:} Please classify the medical multiple choice question into one of the following clinical specialties: *Emergency medicine*, *Allergist*, *Anaesthetics*, *Cardiology*, *Child psychiatry*, *Clinical biology*, *Clinical chemistry*, *Clinical microbiology*, *Clinical neurophysiology*, *Craniofacial surgery*, *Dermatology*, *Endocrinology*, *Family and General Medicine*, *Gastroenterologic surgery*, *Gastroenterology*, *General Practice*, *General surgery*, *Geriatrics*, *Hematology*, *Immunology*, *Infectious diseases*, *Internal medicine*, *Laboratory medicine*, *Nephrology*, *Neuropsychiatry*, *Neurology*, *Neurosurgery*, *Nuclear medicine*, *Obstetrics and gynecology*, *Occupational medicine*, *Oncology*, *Ophthalmology*, *Oral and maxillofacial surgery*, *Orthopedics*, *Otorhinolaryngology*, *Pediatric surgery*, *Pediatrics*, *Pathology*, *Pharmacology*, *Physical medicine and rehabilitation*, *Plastic surgery*, *Podiatric surgery*, *Preventive medicine*, *Psychiatry*, *Public health*, *Radiation Oncology*, *Radiology*, *Respiratory medicine*, *Rheumatology*, *Stomatology*, *Thoracic surgery*, *Tropical medicine*, *Urology*, *Vascular surgery*, *Venereology*, *Others* \\

\textbf{\#\#\# User}: A 39-year-old woman comes to the physician because of an 8-month history of progressive fatigue, shortness of breath, and palpitations. She has a history of recurrent episodes of joint pain and fever during childhood. She emigrated from India with her parents when she was 10 years old. Cardiac examination shows an opening snap followed by a late diastolic rumble, which is best heard at the fifth intercostal space in the left midclavicular line. This patient is at greatest risk for compression of which of the following structures?
\end{minipage}
\end{center}
\end{tcolorbox}
\end{figure*}

%% file: appendices/prompt-2.tex
\begin{figure*}
\begin{tcolorbox}
\caption{Prompt-2}
\label{fig:gpt_prompts_2}
\vspace{0.3cm}
\begin{center}
\begin{minipage}{0.95\textwidth}
\scriptsize
\textbf{\#\#\# System:} You are medical student taking a multiple choice exam. The knowledge of which of the following clinical specialties is the most helpful to answering the question: *Emergency medicine*, *Allergist*, *Anaesthetics*, *Cardiology*, *Child psychiatry*, *Clinical biology*, *Clinical chemistry*, *Clinical microbiology*, *Clinical neurophysiology*, *Craniofacial surgery*, *Dermatology*, *Endocrinology*, *Family and General Medicine*, *Gastroenterologic surgery*, *Gastroenterology*, *General Practice*, *General surgery*, *Geriatrics*, *Hematology*, *Immunology*, *Infectious diseases*, *Internal medicine*, *Laboratory medicine*, *Nephrology*, *Neuropsychiatry*, *Neurology*, *Neurosurgery*, *Nuclear medicine*, *Obstetrics and gynecology*, *Occupational medicine*, *Oncology*, *Ophthalmology*, *Oral and maxillofacial surgery*, *Orthopedics*, *Otorhinolaryngology*, *Pediatric surgery*, *Pediatrics*, *Pathology*, *Pharmacology*, *Physical medicine and rehabilitation*, *Plastic surgery*, *Podiatric surgery*, *Preventive medicine*, *Psychiatry*, *Public health*, *Radiation Oncology*, *Radiology*, *Respiratory medicine*, *Rheumatology*, *Stomatology*, *Thoracic surgery*, *Tropical medicine*, *Urology*, *Vascular surgery*, *Venereology*, *Others* \\

Here are some examples:

Question: A 62-year-old woman presents for a regular check-up. She complains of lightheadedness and palpitations which occur episodically. Past medical history is significant for a myocardial infarction 6 months ago and NYHA class II chronic heart failure. She also was diagnosed with grade I arterial hypertension 4 years ago. Current medications are aspirin 81 mg, atorvastatin 10 mg, enalapril 10 mg, and metoprolol 200 mg daily. Her vital signs are a blood pressure of 135/90 mm Hg, a heart rate of 125/min, a respiratory rate of 14/min, and a temperature of 36.5°C (97.7°F). Cardiopulmonary examination is significant for irregular heart rhythm and decreased S1 intensity. ECG is obtained and is shown in the picture (see image). Echocardiography shows a left ventricular ejection fraction of 39\%. Which of the following drugs is the best choice for rate control in this patient?

Answer: Cardiology

Question: A 68-year-old man comes to the physician because of recurrent episodes of nausea and abdominal discomfort for the past 4 months. The discomfort is located in the upper abdomen and sometimes occurs after eating, especially after a big meal. He has tried to go for a walk after dinner to help with digestion, but his complaints have only increased. For the past 3 weeks he has also had symptoms while climbing the stairs to his apartment. He has type 2 diabetes mellitus, hypertension, and stage 2 peripheral arterial disease. He has smoked one pack of cigarettes daily for the past 45 years. He drinks one to two beers daily and occasionally more on weekends. His current medications include metformin, enalapril, and aspirin. He is 168 cm (5 ft 6 in) tall and weighs 126 kg (278 lb); BMI is 45 kg/m2. His temperature is 36.4°C (97.5°F), pulse is 78/min, and blood pressure is 148/86 mm Hg. On physical examination, the abdomen is soft and nontender with no organomegaly. Foot pulses are absent bilaterally. An ECG shows no abnormalities. Which of the following is the most appropriate next step in diagnosis?

Answer: Gastroenterology

Question: A 6-year-old male who recently immigrated to the United States from Asia is admitted to the hospital with dyspnea. Physical exam reveals a gray pseudomembrane in the patient's oropharynx along with lymphadenopathy. The patient develops myocarditis and expires on hospital day 5. Which of the following would have prevented this patient's presentation and decline?

Answer: Infectious diseases

Question: A 35-year-old woman with a history of Crohn disease presents for a follow-up appointment. She says that lately, she has started to notice difficulty walking. She says that some of her friends have joked that she appears to be walking as if she was drunk. Past medical history is significant for Crohn disease diagnosed 2 years ago, managed with natalizumab for the past year because her intestinal symptoms have become severe and unresponsive to other therapies. On physical examination, there is gait and limb ataxia present. Strength is 4/5 in the right upper limb. A T1/T2 MRI of the brain is ordered and is shown. Which of the following is the most likely diagnosis?

Answer: Neurology

Question: A 23-year-old G1 at 10 weeks gestation based on her last menstrual period is brought to the emergency department by her husband due to sudden vaginal bleeding. She says that she has mild lower abdominal cramps and is feeling dizzy and weak. Her blood pressure is 100/60 mm Hg, the pulse is 100/min, and the respiration rate is 15/min. She says that she has had light spotting over the last 3 days, but today the bleeding increased markedly and she also noticed the passage of clots. She says that she has changed three pads since the morning. She has also noticed that the nausea she was experiencing over the past few days has subsided. The physician examines her and notes that the cervical os is open and blood is pooling in the vagina. Products of conception can be visualized in the os. The patient is prepared for a suction curettage. Which of the following is the most likely cause for the pregnancy loss?

Answer: Obstetrics and gynecology

Question: An 8-month-old boy is brought to a medical office by his mother. The mother states that the boy has been very fussy and has not been feeding recently. The mother thinks the baby has been gaining weight despite not feeding well. The boy was delivered vaginally at 39 weeks gestation without complications. On physical examination, the boy is noted to be crying in his mother’s arms. There is no evidence of cyanosis, and the cardiac examination is within normal limits. The crying intensifies when the abdomen is palpated. The abdomen is distended with tympany in the left lower quadrant. You suspect a condition caused by the failure of specialized cells to migrate. What is the most likely diagnosis?

Answer: Pediatrics \\

\textbf{\#\#\# User:} A 39-year-old woman comes to the physician because of an 8-month history of progressive fatigue, shortness of breath, and palpitations. She has a history of recurrent episodes of joint pain and fever during childhood. She emigrated from India with her parents when she was 10 years old. Cardiac examination shows an opening snap followed by a late diastolic rumble, which is best heard at the fifth intercostal space in the left midclavicular line. This patient is at greatest risk for compression of which of the following structures?
\end{minipage}
\end{center}
\end{tcolorbox}
\end{figure*}

%% file: appendices/prompt-3.tex
\begin{figure*}
\begin{tcolorbox}
\caption{Prompt-3}
\label{fig:gpt_prompts_3}
\vspace{0.3cm}
\begin{center}
\begin{minipage}{0.95\textwidth}
\scriptsize
\textbf{\#\#\# System:} Please classify the medical multiple choice question into one of the following clinical specialties: *Emergency medicine*, *Allergist*, *Anaesthetics*, *Cardiology*, *Child psychiatry*, *Clinical biology*, *Clinical chemistry*, *Clinical microbiology*, *Clinical neurophysiology*, *Craniofacial surgery*, *Dermatology*, *Endocrinology*, *Family and General Medicine*, *Gastroenterologic surgery*, *Gastroenterology*, *General Practice*, *General surgery*, *Geriatrics*, *Hematology*, *Immunology*, *Infectious diseases*, *Internal medicine*, *Laboratory medicine*, *Nephrology*, *Neuropsychiatry*, *Neurology*, *Neurosurgery*, *Nuclear medicine*, *Obstetrics and gynecology*, *Occupational medicine*, *Oncology*, *Ophthalmology*, *Oral and maxillofacial surgery*, *Orthopedics*, *Otorhinolaryngology*, *Pediatric surgery*, *Pediatrics*, *Pathology*, *Pharmacology*, *Physical medicine and rehabilitation*, *Plastic surgery*, *Podiatric surgery*, *Preventive medicine*, *Psychiatry*, *Public health*, *Radiation Oncology*, *Radiology*, *Respiratory medicine*, *Rheumatology*, *Stomatology*, *Thoracic surgery*, *Tropical medicine*, *Urology*, *Vascular surgery*, *Venereology*, *Others* \\

Here are some examples:

Question: A 62-year-old woman presents for a regular check-up. She complains of lightheadedness and palpitations which occur episodically. Past medical history is significant for a myocardial infarction 6 months ago and NYHA class II chronic heart failure. She also was diagnosed with grade I arterial hypertension 4 years ago. Current medications are aspirin 81 mg, atorvastatin 10 mg, enalapril 10 mg, and metoprolol 200 mg daily. Her vital signs are a blood pressure of 135/90 mm Hg, a heart rate of 125/min, a respiratory rate of 14/min, and a temperature of 36.5°C (97.7°F). Cardiopulmonary examination is significant for irregular heart rhythm and decreased S1 intensity. ECG is obtained and is shown in the picture (see image). Echocardiography shows a left ventricular ejection fraction of 39\%. Which of the following drugs is the best choice for rate control in this patient?

Answer: Cardiology

Question: A 68-year-old man comes to the physician because of recurrent episodes of nausea and abdominal discomfort for the past 4 months. The discomfort is located in the upper abdomen and sometimes occurs after eating, especially after a big meal. He has tried to go for a walk after dinner to help with digestion, but his complaints have only increased. For the past 3 weeks he has also had symptoms while climbing the stairs to his apartment. He has type 2 diabetes mellitus, hypertension, and stage 2 peripheral arterial disease. He has smoked one pack of cigarettes daily for the past 45 years. He drinks one to two beers daily and occasionally more on weekends. His current medications include metformin, enalapril, and aspirin. He is 168 cm (5 ft 6 in) tall and weighs 126 kg (278 lb); BMI is 45 kg/m2. His temperature is 36.4°C (97.5°F), pulse is 78/min, and blood pressure is 148/86 mm Hg. On physical examination, the abdomen is soft and nontender with no organomegaly. Foot pulses are absent bilaterally. An ECG shows no abnormalities. Which of the following is the most appropriate next step in diagnosis?

Answer: Gastroenterology

Question: A 6-year-old male who recently immigrated to the United States from Asia is admitted to the hospital with dyspnea. Physical exam reveals a gray pseudomembrane in the patient's oropharynx along with lymphadenopathy. The patient develops myocarditis and expires on hospital day 5. Which of the following would have prevented this patient's presentation and decline?

Answer: Infectious diseases

Question: A 35-year-old woman with a history of Crohn disease presents for a follow-up appointment. She says that lately, she has started to notice difficulty walking. She says that some of her friends have joked that she appears to be walking as if she was drunk. Past medical history is significant for Crohn disease diagnosed 2 years ago, managed with natalizumab for the past year because her intestinal symptoms have become severe and unresponsive to other therapies. On physical examination, there is gait and limb ataxia present. Strength is 4/5 in the right upper limb. A T1/T2 MRI of the brain is ordered and is shown. Which of the following is the most likely diagnosis?

Answer: Neurology

Question: A 23-year-old G1 at 10 weeks gestation based on her last menstrual period is brought to the emergency department by her husband due to sudden vaginal bleeding. She says that she has mild lower abdominal cramps and is feeling dizzy and weak. Her blood pressure is 100/60 mm Hg, the pulse is 100/min, and the respiration rate is 15/min. She says that she has had light spotting over the last 3 days, but today the bleeding increased markedly and she also noticed the passage of clots. She says that she has changed three pads since the morning. She has also noticed that the nausea she was experiencing over the past few days has subsided. The physician examines her and notes that the cervical os is open and blood is pooling in the vagina. Products of conception can be visualized in the os. The patient is prepared for a suction curettage. Which of the following is the most likely cause for the pregnancy loss?

Answer: Obstetrics and gynecology

Question: An 8-month-old boy is brought to a medical office by his mother. The mother states that the boy has been very fussy and has not been feeding recently. The mother thinks the baby has been gaining weight despite not feeding well. The boy was delivered vaginally at 39 weeks gestation without complications. On physical examination, the boy is noted to be crying in his mother’s arms. There is no evidence of cyanosis, and the cardiac examination is within normal limits. The crying intensifies when the abdomen is palpated. The abdomen is distended with tympany in the left lower quadrant. You suspect a condition caused by the failure of specialized cells to migrate. What is the most likely diagnosis?

Answer: Pediatrics \\

\textbf{\#\#\# User:} A 39-year-old woman comes to the physician because of an 8-month history of progressive fatigue, shortness of breath, and palpitations. She has a history of recurrent episodes of joint pain and fever during childhood. She emigrated from India with her parents when she was 10 years old. Cardiac examination shows an opening snap followed by a late diastolic rumble, which is best heard at the fifth intercostal space in the left midclavicular line. This patient is at greatest risk for compression of which of the following structures?
\end{minipage}
\end{center}
\end{tcolorbox}
\end{figure*}

%% file: appendices/prompt-4.tex
\begin{figure*}
\begin{tcolorbox}
\caption{Prompt-4}
\label{fig:gpt_prompts_4}
\vspace{0.3cm}
\begin{center}
\begin{minipage}{0.95\textwidth}
\scriptsize
\textbf{\#\#\# System:} Please classify the medical multiple choice question into one of the clinical specialties. \\

Here are some examples:

Question: A 62-year-old woman presents for a regular check-up. She complains of lightheadedness and palpitations which occur episodically. Past medical history is significant for a myocardial infarction 6 months ago and NYHA class II chronic heart failure. She also was diagnosed with grade I arterial hypertension 4 years ago. Current medications are aspirin 81 mg, atorvastatin 10 mg, enalapril 10 mg, and metoprolol 200 mg daily. Her vital signs are a blood pressure of 135/90 mm Hg, a heart rate of 125/min, a respiratory rate of 14/min, and a temperature of 36.5°C (97.7°F). Cardiopulmonary examination is significant for irregular heart rhythm and decreased S1 intensity. ECG is obtained and is shown in the picture (see image). Echocardiography shows a left ventricular ejection fraction of 39\%. Which of the following drugs is the best choice for rate control in this patient?

Answer: Cardiology

Question: A 68-year-old man comes to the physician because of recurrent episodes of nausea and abdominal discomfort for the past 4 months. The discomfort is located in the upper abdomen and sometimes occurs after eating, especially after a big meal. He has tried to go for a walk after dinner to help with digestion, but his complaints have only increased. For the past 3 weeks he has also had symptoms while climbing the stairs to his apartment. He has type 2 diabetes mellitus, hypertension, and stage 2 peripheral arterial disease. He has smoked one pack of cigarettes daily for the past 45 years. He drinks one to two beers daily and occasionally more on weekends. His current medications include metformin, enalapril, and aspirin. He is 168 cm (5 ft 6 in) tall and weighs 126 kg (278 lb); BMI is 45 kg/m2. His temperature is 36.4°C (97.5°F), pulse is 78/min, and blood pressure is 148/86 mm Hg. On physical examination, the abdomen is soft and nontender with no organomegaly. Foot pulses are absent bilaterally. An ECG shows no abnormalities. Which of the following is the most appropriate next step in diagnosis?

Answer: Gastroenterology

Question: A 6-year-old male who recently immigrated to the United States from Asia is admitted to the hospital with dyspnea. Physical exam reveals a gray pseudomembrane in the patient's oropharynx along with lymphadenopathy. The patient develops myocarditis and expires on hospital day 5. Which of the following would have prevented this patient's presentation and decline?

Answer: Infectious diseases

Question: A 35-year-old woman with a history of Crohn disease presents for a follow-up appointment. She says that lately, she has started to notice difficulty walking. She says that some of her friends have joked that she appears to be walking as if she was drunk. Past medical history is significant for Crohn disease diagnosed 2 years ago, managed with natalizumab for the past year because her intestinal symptoms have become severe and unresponsive to other therapies. On physical examination, there is gait and limb ataxia present. Strength is 4/5 in the right upper limb. A T1/T2 MRI of the brain is ordered and is shown. Which of the following is the most likely diagnosis?

Answer: Neurology

Question: A 23-year-old G1 at 10 weeks gestation based on her last menstrual period is brought to the emergency department by her husband due to sudden vaginal bleeding. She says that she has mild lower abdominal cramps and is feeling dizzy and weak. Her blood pressure is 100/60 mm Hg, the pulse is 100/min, and the respiration rate is 15/min. She says that she has had light spotting over the last 3 days, but today the bleeding increased markedly and she also noticed the passage of clots. She says that she has changed three pads since the morning. She has also noticed that the nausea she was experiencing over the past few days has subsided. The physician examines her and notes that the cervical os is open and blood is pooling in the vagina. Products of conception can be visualized in the os. The patient is prepared for a suction curettage. Which of the following is the most likely cause for the pregnancy loss?

Answer: Obstetrics and gynecology

Question: An 8-month-old boy is brought to a medical office by his mother. The mother states that the boy has been very fussy and has not been feeding recently. The mother thinks the baby has been gaining weight despite not feeding well. The boy was delivered vaginally at 39 weeks gestation without complications. On physical examination, the boy is noted to be crying in his mother’s arms. There is no evidence of cyanosis, and the cardiac examination is within normal limits. The crying intensifies when the abdomen is palpated. The abdomen is distended with tympany in the left lower quadrant. You suspect a condition caused by the failure of specialized cells to migrate. What is the most likely diagnosis?

Answer: Pediatrics \\

\textbf{\#\#\# User:} A 39-year-old woman comes to the physician because of an 8-month history of progressive fatigue, shortness of breath, and palpitations. She has a history of recurrent episodes of joint pain and fever during childhood. She emigrated from India with her parents when she was 10 years old. Cardiac examination shows an opening snap followed by a late diastolic rumble, which is best heard at the fifth intercostal space in the left midclavicular line. This patient is at greatest risk for compression of which of the following structures? 

Please classify the medical multiple choice question into one of the following clinical specialties: *Emergency medicine*, *Allergist*, *Anaesthetics*, *Cardiology*, *Child psychiatry*, *Clinical biology*, *Clinical chemistry*, *Clinical microbiology*, *Clinical neurophysiology*, *Craniofacial surgery*, *Dermatology*, *Endocrinology*, *Family and General Medicine*, *Gastroenterologic surgery*, *Gastroenterology*, *General Practice*, *General surgery*, *Geriatrics*, *Hematology*, *Immunology*, *Infectious diseases*, *Internal medicine*, *Laboratory medicine*, *Nephrology*, *Neuropsychiatry*, *Neurology*, *Neurosurgery*, *Nuclear medicine*, *Obstetrics and gynecology*, *Occupational medicine*, *Oncology*, *Ophthalmology*, *Oral and maxillofacial surgery*, *Orthopedics*, *Otorhinolaryngology*, *Pediatric surgery*, *Pediatrics*, *Pathology*, *Pharmacology*, *Physical medicine and rehabilitation*, *Plastic surgery*, *Podiatric surgery*, *Preventive medicine*, *Psychiatry*, *Public health*, *Radiation Oncology*, *Radiology*, *Respiratory medicine*, *Rheumatology*, *Stomatology*, *Thoracic surgery*, *Tropical medicine*, *Urology*, *Vascular surgery*, *Venereology*, *Others* \\
\end{minipage}
\end{center}
\end{tcolorbox}
\end{figure*}

%% file: appendices/prompt-5.tex
\begin{figure*}
\begin{tcolorbox}
\caption{Prompt-5}
\label{fig:gpt_prompts_5}
\vspace{0.3cm}
\begin{center}
\begin{minipage}{0.95\textwidth}
\scriptsize
\textbf{\#\#\# System:} Please classify the medical multiple choice question into one of the following clinical specialties: *Emergency medicine*, *Allergist*, *Anaesthetics*, *Cardiology*, *Child psychiatry*, *Clinical biology*, *Clinical chemistry*, *Clinical microbiology*, *Clinical neurophysiology*, *Craniofacial surgery*, *Dermatology*, *Endocrinology*, *Family and General Medicine*, *Gastroenterologic surgery*, *Gastroenterology*, *General Practice*, *General surgery*, *Geriatrics*, *Hematology*, *Immunology*, *Infectious diseases*, *Internal medicine*, *Laboratory medicine*, *Nephrology*, *Neuropsychiatry*, *Neurology*, *Neurosurgery*, *Nuclear medicine*, *Obstetrics and gynecology*, *Occupational medicine*, *Oncology*, *Ophthalmology*, *Oral and maxillofacial surgery*, *Orthopedics*, *Otorhinolaryngology*, *Pediatric surgery*, *Pediatrics*, *Pathology*, *Pharmacology*, *Physical medicine and rehabilitation*, *Plastic surgery*, *Podiatric surgery*, *Preventive medicine*, *Psychiatry*, *Public health*, *Radiation Oncology*, *Radiology*, *Respiratory medicine*, *Rheumatology*, *Stomatology*, *Thoracic surgery*, *Tropical medicine*, *Urology*, *Vascular surgery*, *Venereology*, *Others* \\

Here are some examples:

Question: A 62-year-old woman presents for a regular check-up. She complains of lightheadedness and palpitations which occur episodically. Past medical history is significant for a myocardial infarction 6 months ago and NYHA class II chronic heart failure. She also was diagnosed with grade I arterial hypertension 4 years ago. Current medications are aspirin 81 mg, atorvastatin 10 mg, enalapril 10 mg, and metoprolol 200 mg daily. Her vital signs are a blood pressure of 135/90 mm Hg, a heart rate of 125/min, a respiratory rate of 14/min, and a temperature of 36.5°C (97.7°F). Cardiopulmonary examination is significant for irregular heart rhythm and decreased S1 intensity. ECG is obtained and is shown in the picture (see image). Echocardiography shows a left ventricular ejection fraction of 39\%. Which of the following drugs is the best choice for rate control in this patient?

Answer: Cardiology

Question: A 68-year-old man comes to the physician because of recurrent episodes of nausea and abdominal discomfort for the past 4 months. The discomfort is located in the upper abdomen and sometimes occurs after eating, especially after a big meal. He has tried to go for a walk after dinner to help with digestion, but his complaints have only increased. For the past 3 weeks he has also had symptoms while climbing the stairs to his apartment. He has type 2 diabetes mellitus, hypertension, and stage 2 peripheral arterial disease. He has smoked one pack of cigarettes daily for the past 45 years. He drinks one to two beers daily and occasionally more on weekends. His current medications include metformin, enalapril, and aspirin. He is 168 cm (5 ft 6 in) tall and weighs 126 kg (278 lb); BMI is 45 kg/m2. His temperature is 36.4°C (97.5°F), pulse is 78/min, and blood pressure is 148/86 mm Hg. On physical examination, the abdomen is soft and nontender with no organomegaly. Foot pulses are absent bilaterally. An ECG shows no abnormalities. Which of the following is the most appropriate next step in diagnosis?

Answer: Gastroenterology

Question: A 6-year-old male who recently immigrated to the United States from Asia is admitted to the hospital with dyspnea. Physical exam reveals a gray pseudomembrane in the patient's oropharynx along with lymphadenopathy. The patient develops myocarditis and expires on hospital day 5. Which of the following would have prevented this patient's presentation and decline?

Answer: Infectious diseases

Question: A 35-year-old woman with a history of Crohn disease presents for a follow-up appointment. She says that lately, she has started to notice difficulty walking. She says that some of her friends have joked that she appears to be walking as if she was drunk. Past medical history is significant for Crohn disease diagnosed 2 years ago, managed with natalizumab for the past year because her intestinal symptoms have become severe and unresponsive to other therapies. On physical examination, there is gait and limb ataxia present. Strength is 4/5 in the right upper limb. A T1/T2 MRI of the brain is ordered and is shown. Which of the following is the most likely diagnosis?

Answer: Neurology

Question: A 23-year-old G1 at 10 weeks gestation based on her last menstrual period is brought to the emergency department by her husband due to sudden vaginal bleeding. She says that she has mild lower abdominal cramps and is feeling dizzy and weak. Her blood pressure is 100/60 mm Hg, the pulse is 100/min, and the respiration rate is 15/min. She says that she has had light spotting over the last 3 days, but today the bleeding increased markedly and she also noticed the passage of clots. She says that she has changed three pads since the morning. She has also noticed that the nausea she was experiencing over the past few days has subsided. The physician examines her and notes that the cervical os is open and blood is pooling in the vagina. Products of conception can be visualized in the os. The patient is prepared for a suction curettage. Which of the following is the most likely cause for the pregnancy loss?

Answer: Obstetrics and gynecology

Question: An 8-month-old boy is brought to a medical office by his mother. The mother states that the boy has been very fussy and has not been feeding recently. The mother thinks the baby has been gaining weight despite not feeding well. The boy was delivered vaginally at 39 weeks gestation without complications. On physical examination, the boy is noted to be crying in his mother’s arms. There is no evidence of cyanosis, and the cardiac examination is within normal limits. The crying intensifies when the abdomen is palpated. The abdomen is distended with tympany in the left lower quadrant. You suspect a condition caused by the failure of specialized cells to migrate. What is the most likely diagnosis?

Answer: Pediatrics \\

\textbf{\#\#\# User:} Question: A 39-year-old woman comes to the physician because of an 8-month history of progressive fatigue, shortness of breath, and palpitations. She has a history of recurrent episodes of joint pain and fever during childhood. She emigrated from India with her parents when she was 10 years old. Cardiac examination shows an opening snap followed by a late diastolic rumble, which is best heard at the fifth intercostal space in the left midclavicular line. This patient is at greatest risk for compression of which of the following structures?

Answer: \\
\end{minipage}
\end{center}
\end{tcolorbox}
\end{figure*}

%% file: tables/equal_train.tex
\begin{table*}[t]
\centering
\scalebox{0.75}{
\begin{tabular}{>{\centering\arraybackslash}m{1em} >{\raggedright\arraybackslash}m{7em} >{\centering\arraybackslash}m{4em} >{\centering\arraybackslash}m{4em} >{\centering\arraybackslash}m{4em} >{\centering\arraybackslash}m{4em} >{\centering\arraybackslash}m{4em} >{\centering\arraybackslash}m{4em} >{\centering\arraybackslash}m{4em}}

& \textbf{Test Sets} & Cardio & Gastro & Infect & Neuro & Obstetrics & Pediatrics & avg. \\
\cmidrule{2-9}
\multirow{8}{*}{\STAB{\rotatebox[origin=c]{90}{Train Sets}}} & Llama-3.1 & 64.0 & 67.5 & 65.3 & 66.8 & 65.0 & 64.5 & 65.5 \\
\cmidrule{3-9}
& Cardio & \underline{\textbf{71.8}} & \textbf{77.4} & 72.3 & 76.1 & \textbf{69.9} & 71.0 & 73.0 \\
& Gastro & 69.6 & \underline{77.1} & 72.4 & 77.2 & 69.1 & 71.9 & 72.8 \\
& Infect & 68.8 & 74.5 & \underline{73.3} & \textbf{77.4} & 69.8 & 71.3 & 72.4 \\
& Neuro & 70.8 & 77.1 & \textbf{73.4} & \underline{76.9} & 68.6 & \textbf{72.1} & 73.0 \\
& Obstetrics & 69.8 & 75.0 & 72.9 & 75.4 & \underline{68.7} & 69.4 & 71.9 \\
& Pediatrics & 68.9 & 75.3 & 72.8 & 77.0 & 69.1 & \underline{70.4} & 72.2 \\
\cmidrule{2-9}
\end{tabular}}
\caption{Cross specialty evaluation results fine-tuned on training data with equal sizes.}
\label{tab:equal_train}
\end{table*}

%% file: tables/bootstrap.tex
\begin{table*}[t]
\centering
\scalebox{0.75}{
\begin{tabular}{>{\centering\arraybackslash}m{1em} >{\raggedright\arraybackslash}m{7em} >{\centering\arraybackslash}m{4em} >{\centering\arraybackslash}m{4em} >{\centering\arraybackslash}m{4em} >{\centering\arraybackslash}m{4em} >{\centering\arraybackslash}m{4em} >{\centering\arraybackslash}m{4em} >{\centering\arraybackslash}m{4em}}

& \textbf{Test Sets} & Cardio & Gastro & Infect & Neuro & Obstetrics & Pediatrics & avg. \\
\cmidrule{2-9}
\multirow{8}{*}{\STAB{\rotatebox[origin=c]{90}{Train Sets}}} & Llama-3.1 & 70.1 & 72.3 & 71.0 & 72.8 & 70.2 & 69.4 & 71.0 \\
\cmidrule{3-9}
& Cardio & \underline{\textbf{79.7}} & 75.4 & 72.3 & \textbf{77.6} & \textbf{73.1} & 71.1 & 73.6 \\
& Gastro & 70.0 & \underline{\textbf{77.4}} & 72.2 & 77.2 & 73.0 & \textbf{72.7} & 73.6 \\
& Infect & 70.4 & 73.8 & \underline{72.7} & 77.3 & 71.0 & 70.3 & 72.6 \\
& Neuro & 72.0 & 76.0 & 72.7 & \underline{77.3} & 72.0 & 71.3 & 73.4 \\
& Obstetrics & 70.3 & 75.4 & 71.9 & 75.7 & \underline{72.5} & 69.5 & 72.4 \\
& Pediatrics & 71.2 & 74.8 & \textbf{72.9} & 76.4 & 71.4 & \underline{70.8} & 72.8 \\
\cmidrule{3-9}
& Combined & 63.1 & 81.4 & 77.5 & 78.2 & 73.3 & 76.6 & 76.7 \\
\cmidrule{2-9}
\end{tabular}}
\caption{Cross specialty evaluation results of 20 resamplings of the test sets.}
\label{tab:bootstrap}
\end{table*}